\documentclass[10pt,journal,letterpaper,twoside]{IEEEtran}
\usepackage[color=red!60]{todonotes}
\usepackage[caption=false]{subfig}

\usepackage{cite}

\usepackage{graphicx}
\usepackage{amsmath}
\usepackage{multirow}
\usepackage{amssymb} 
\usepackage{bm} 
\usepackage{algorithm}
\usepackage{algpseudocode}

\usepackage{booktabs}
\usepackage{tabularx}

\usepackage{hyperref}

\newcommand{\x}[0]{\ensuremath{\boldsymbol{x}}}

\newcommand{\nwindows}[0]{\ensuremath{W}}
\newcommand{\windowind}[0]{\ensuremath{w}}

\newcommand{\nclasses}[0]{\ensuremath{C}}
\newcommand{\classind}[0]{\ensuremath{c}}

\newcommand{\ntrials}[0]{\ensuremath{R}}
\newcommand{\trialind}[0]{\ensuremath{j}}

\newcommand{\nmodels}[0]{\ensuremath{M}}
\newcommand{\modelind}[0]{\ensuremath{m}}

\newcommand{\exind}[0]{\ensuremath{i}}

\newcommand{\npred}[0]{\ensuremath{O}}
\newcommand{\predind}[0]{\ensuremath{k}}

\newcommand{\kl}[2]{\ensuremath{\text{KL}(#1||#2)}}

\usepackage{scalerel}
\usepackage{tikz}
\usetikzlibrary{svg.path}
\definecolor{orcidlogocol}{HTML}{A6CE39}
\tikzset{
    orcidlogo/.pic={
        \fill[orcidlogocol] svg{M256,128c0,70.7-57.3,128-128,128C57.3,256,0,198.7,0,128C0,57.3,57.3,0,128,0C198.7,0,256,57.3,256,128z};
        \fill[white] svg{M86.3,186.2H70.9V79.1h15.4v48.4V186.2z}
        svg{M108.9,79.1h41.6c39.6,0,57,28.3,57,53.6c0,27.5-21.5,53.6-56.8,53.6h-41.8V79.1z M124.3,172.4h24.5c34.9,0,42.9-26.5,42.9-39.7c0-21.5-13.7-39.7-43.7-39.7h-23.7V172.4z}
        svg{M88.7,56.8c0,5.5-4.5,10.1-10.1,10.1c-5.6,0-10.1-4.6-10.1-10.1c0-5.6,4.5-10.1,10.1-10.1C84.2,46.7,88.7,51.3,88.7,56.8z};
    }
}

\newcommand\orcidicon[1]{\href{https://orcid.org/#1}{\mbox{\scalerel*{
                \begin{tikzpicture}[yscale=-1,transform shape]
                \pic{orcidlogo};
                \end{tikzpicture}
            }{|}}}}

\begin{document}

%
\title{Ensembles of Spiking Neural Networks}
%
%
%

\author{Georgiana~Neculae$^{\textsuperscript{\orcidicon{0000-0001-6541-1589}}}$,
        Oliver~Rhodes$^{\textsuperscript{\orcidicon{0000-0003-1728-2828}}}$and
        Gavin~Brown$^{\textsuperscript{\orcidicon{0000-0003-2261-9018}}}$
}
\maketitle

\begin{abstract}

This paper demonstrates how to construct ensembles of spiking neural networks producing state-of-the-art results, achieving classification accuracies of 98.71\%, 100.0\%, and 99.09\%, on the MNIST, NMNIST and DVS Gesture datasets respectively. Furthermore, this performance is achieved using simplified individual models, with ensembles containing less than 50\% of the parameters of published reference models. We provide comprehensive exploration on the effect of spike train interpretation methods, and derive the theoretical methodology for combining model predictions such that performance improvements are guaranteed for spiking ensembles. For this, we formalize spiking neural networks as GLM predictors, identifying a suitable representation for their target domain. Further, we show how the diversity of our spiking ensembles can be measured using the Ambiguity Decomposition. The work demonstrates how ensembling can overcome the challenges of producing individual SNN models which can compete with traditional deep neural networks, and creates systems with fewer trainable parameters and smaller memory footprints, opening the door to low-power edge applications, e.g. implemented on neuromorphic hardware.
\end{abstract}


%
\IEEEpeerreviewmaketitle

\section{Introduction}

%
%
%
%
%
%

\IEEEPARstart{E}{nsemble} systems have been shown to be successful both in machine learning and in the nervous system. When applied to machine learning models, ensembles commonly attain state-of-the-art results, outperforming even the most complex systems \cite{he2016deep,touvron2019fixing}. In the brain, groups of neurons commonly collaborate to produce complex behaviors that would not be possible with a single neuron or a single neural circuit \cite{averbeck2006neural,averbeck2004coding,kopell2011neuronal,o2006biologically,pillow2008spatio}.

Spiking neural networks (SNNs) are driven by advances in neuroscience \cite{gerstner2002spiking,hodgkin1952quantitative}, drawing inspiration from biological neurons which communicate information through temporal events known as `spikes' \cite{liu2014event}, and sharing many of their desirable properties (e.g.~computational and energy efficiency) \cite{benosman2013event}. In spite of their many similarities to biological neurons, SNNs are yet to challenge human performance, and require specialised training to compete with conventional neural networks on the same problems \cite{mozafari2018combining}. However, despite these current limitations, SNNs also offer huge potential, with next-generation `neuromorphic' hardware opening up ultra-low power execution of SNN-based AI algorithms \cite{davies2021advancing}. Furthermore, the spike-based communication of spiking neurons makes them the ideal candidate to process event-based data streams such as from digital vision sensors \cite{amir2017low}, offering low-latency intelligent systems for always-on edge computing.  

Training SNNs represents a significant challenge, as the `spike' discontinuity prevents direct application of gradient-based optimization approaches \cite{Goodfellow-et-al-2016,kelley1960gradient,schmidhuber2015deep, paugam2012computing}. One solution to this problem is to approximate the derivative about the spike using a surrogate function, removing the discontinuity and enabling error-back-propagation-based training. This approach has seen significant recent success, with a number of research groups developing implementations exploring a variety of pseudo-derivative functions \cite{shrestha2018slayer,neftci2019surrogate}. An alternative approach to training SNNs is to harness neuroscience-inspired methods, e.g. utilising online learning mechanisms such as spike timing dependent plasticity (STDP) \cite{diehl2015unsupervised}, but these too come with their limitations (e.g.~designed for shallow architectures). While both approaches can achieve promising results, performance can be expensive, requiring models with large numbers of trainable parameters, and high numbers of training iterations. A target of this work is the use of ensembling to achieve high accuracy predictions using simplified models, reducing the number of trainable parameters and hence the cost of training. While this simplification will inevitably compromise performance of individual models, we investigate whether inclusion of multiple smaller models within an ensemble can recover any lost accuracy. 

An arguably more significant challenge in training SNNs is that outputs of spiking neurons are uninterpretable in their standard format -- collections of times at which the neuron has activated, commonly referred to as spike trains \cite{perkel1968neural}. Lossy interpretation methods (also known as decoding methods) are required to make human-understandable predictions. The main aspect that differentiates these interpretation methods is the quantitative statistical measure of a spike train (e.g.~mean spiking rate) \cite{perkel1968neural} each of them assumes to be carrying the necessary information for decoding the prediction \cite{dayan2001theoretical,diehl2015fast,georgopoulos1986neuronal,gerstner2014neuronal,greschner2006complex,panzeri2001unified,rieke1999spikes,zemel1998probabilistic}. The importance of the interpretation method cannot, however, be understated, as depending on which method is used, a different outcome could be produced. In this work a range of interpretation methods are compared on a variety of problems, highlighting how certain techniques exploit spike-train characteristics to improve prediction accuracy.

\begin{figure}
	\centering
	\hskip-0.5cm
	\includegraphics[width=.475\textwidth]{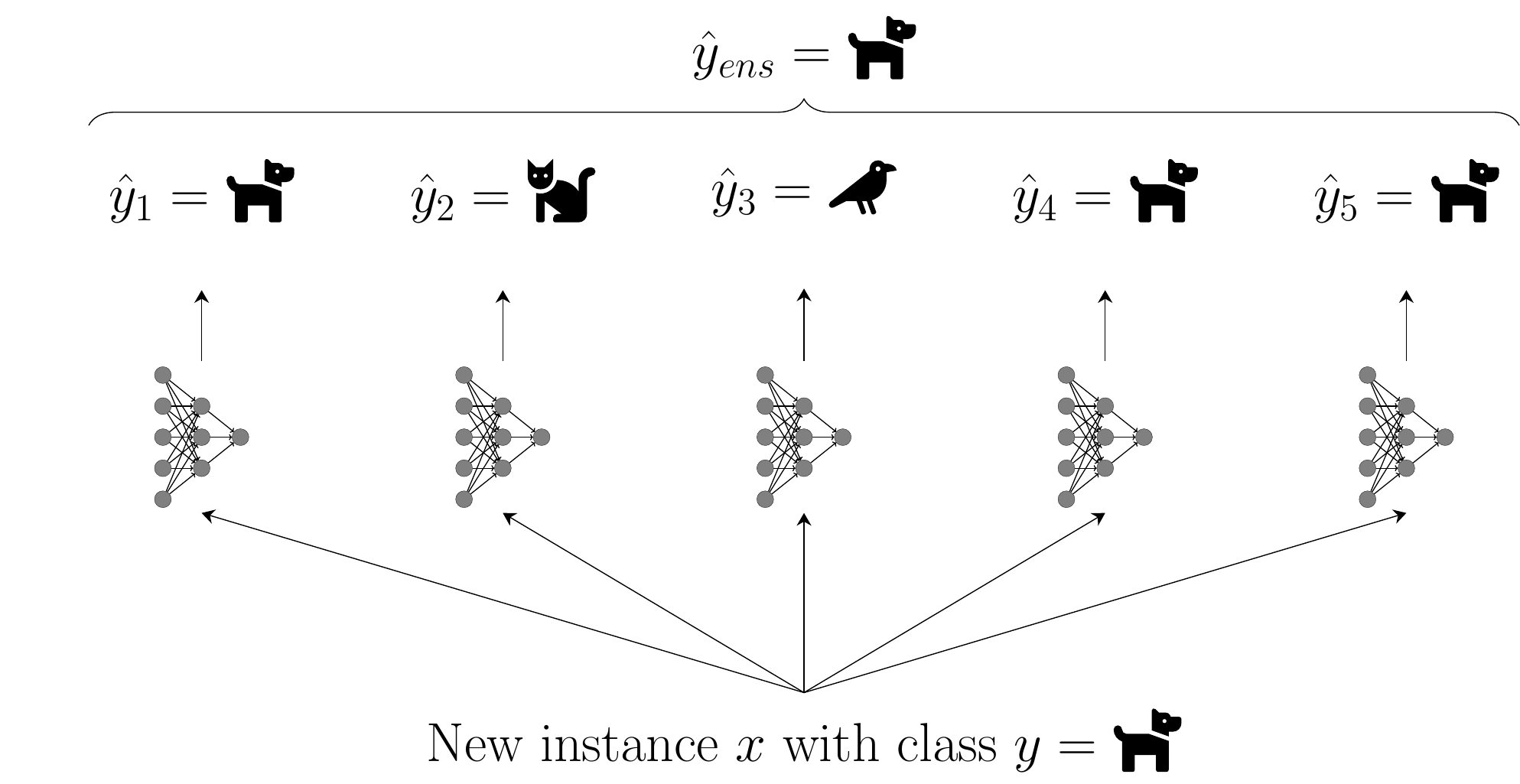}	
	\caption{An example ensemble containing 5 spiking neural networks: the overall ensemble prediction $\hat{y}$ is a combination of the single models.}
	\label{fig:ensemble_diagram}
\end{figure}

\subsection{Related work}
Ensemble systems combine the output of multiple independent models (as seen in Fig.~\ref{fig:ensemble_diagram}), using the notion that a group will typically make a more informed decision than an individual. Ideally, an ensemble system should guarantee an improvement in performance over a single model, otherwise there is no incentive for training additional models. Based on the extensive body of work (in machine learning) on ensembling, two features which have a significant effect on the expected performance of the ensemble system are the combination method, and the diversity of the model predictions \cite{kuncheva2014combining,polikar2006ensemble}. In classical machine learning, certain combination methods \cite{heskes1998selecting,kuncheva2014combining,miller1999critic,webb2019joint} guarantee that the ensemble performance will be higher than the average performance of a single model. The work presented here derives methods to extend this guarantee to SNNs, helping overcome performance limitations introduced through the large variability in SNN predictions (due in part to the representation used, the training process, and the interpretation method). Another property of combination methods with this guarantee is that the ensemble loss can be decomposed in terms of the average error of the individuals, and their average divergence from the ensemble predictions (or disagreement), also known as the Ambiguity Decomposition (henceforth AD) \cite{brown2004diversity,heskes1998selecting,krogh1995neural}.

The presented ensemble methods represent novelty through their exploration of the impact of both the spike train interpretation method, and the ensemble combination method, on overall ensemble performance. 
Kozdon et al.\cite{kozdon2017wide} diverge from the categorization established above, 
combining spike trains by summing individual model spike trains. However, from a rate coding perspective this is equivalent to summing the spike counts of the models. Instead of predicting the class with the highest firing rate, a self organizing map (SOM) is used to measure the similarity of spike trains to class collections established after training. This has the advantage of considering time coding information (depending on the distance function used in the SOM) as well as spiking frequency when making predictions with single models. However, once the spike trains generated by different models are summed, any meaningful spiking patterns are lost at the ensemble level and predictions are again solely based on spike frequency.

In this paper we establish the methodology for ensembling SNNs such that the ensemble is guaranteed to outperform the average single model. We show two ways of formalizing spiking predictions as samples of distributions with an exponential family (Section \ref{sec:spiking_repr}), deriving the combination method and AD for Poisson distributed predictions (Section~\ref{subsec:comb_pred}), and introducing a novel interpretation method that applies principles of ensemble combination methods (Section~\ref{subsec:cfr}). We show in Section \ref{sec:exp} that when ensembles are constructed with one of these approaches, significant increases in accuracy can be produced with much smaller models. We demonstrate this is due to diversity by decomposing the ensemble error using the AD, showing in each case how diversity acts to reduce the average error of the models and gaining insights into the properties of the interpretation methods considered. As the goal is to study the performance of ensembles, rather than push the state of the art for a single SNN, existing methods from the literature are used to train the individual models. Two methods are explored: a bio-inspired unsupervised STDP configuration (see Section~\ref{sec:d_and_c_STDP}); and the supervised gradient-based Spike Layer Reassignment in Time method (SLAYER, see Section~\ref{sec:SLAYER}). Together, these methods represent common approaches to training SNNs in the theoretical neuroscience community, meaning demonstration of the power of ensembling in conjunction with these techniques should appeal to a wide audience. Published results using these methods also exist for a range of datasets, enabling direct comparison of ensembled results against state of the art performance. Our results show that individual models can be simplified  significantly (reductions of $\geq 10\times$ reduction in trainable parameters across all datasets), while improving overall performance when ensembled. This highlights the benefit of ensembles of SNNs, showcasing their impressive performance while exhibiting a smaller memory footprint and reduced training times. 

Following this introduction is an overview of SNNs, together with details of the learning methods used to train each SNN within an ensemble (Section~\ref{sec:spiking_neurons_bck}). Ensembling methods are then discussed, first focusing on interpretation of individual SNN predictions (Section~\ref{sec:spiking_repr}), and then detailing methods for combining predictions within an ensemble (Section~\ref{sec:ensemble_bck}). Finally, experiments are performed evaluating ensemble performance on a range of tasks using a range of neural network architectures (Section~\ref{sec:exp}), followed by summary discussion (Section~\ref{sec:discussion}).

\section{SNN Background \& Methods}
\label{sec:spiking_neurons_bck}

\begin{figure}[!t]
	\centering
	\includegraphics[width=.47\textwidth]{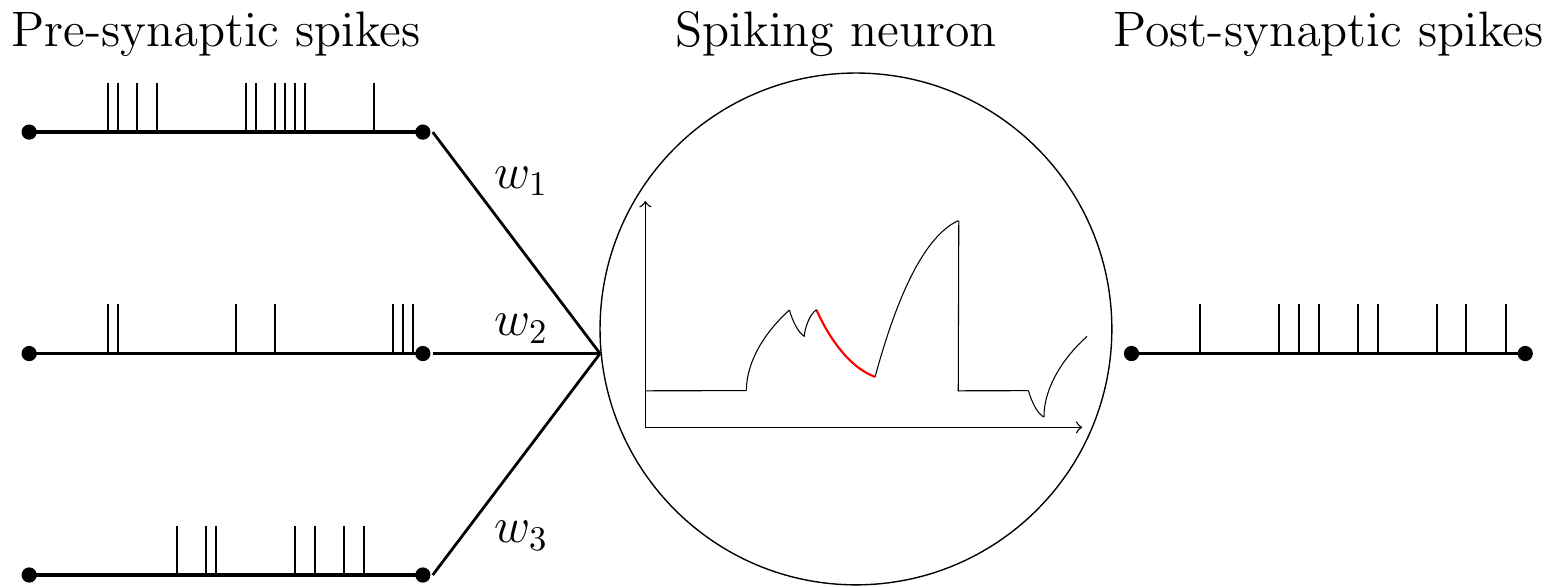}	
	\caption{A spiking neuron receives spikes from three pre-synaptic neurons, which cause it to produce a number of post-synaptic spikes.}
	\label{fig:snn_neuron}
\end{figure}
Spiking neurons incorporate the concept of time during computation, unlike classical neural models \cite{Goodfellow-et-al-2016} (e.g.~ReLu). This is established both through the representation used to encode information (spike trains) and through the neural model (i.e.~the function that describes the effect of input signals on the target neuron). Following the example illustrated in Fig.~\ref{fig:snn_neuron}, a spiking neuron receives as input a set of pre-synaptic spike times (which affect the neuron's activity by an amount proportional to the weight of the synapse connecting the current neuron and the source pre-synaptic neuron), and produces as output a set of post-synaptic spike times \cite{hodgkin1952quantitative,dayan2001theoretical,gerstner2002spiking,liu2014event}.

\subsection{Spiking neuron model}

\begin{figure}[!t]
	\centering
	\includegraphics[width=.47\textwidth]{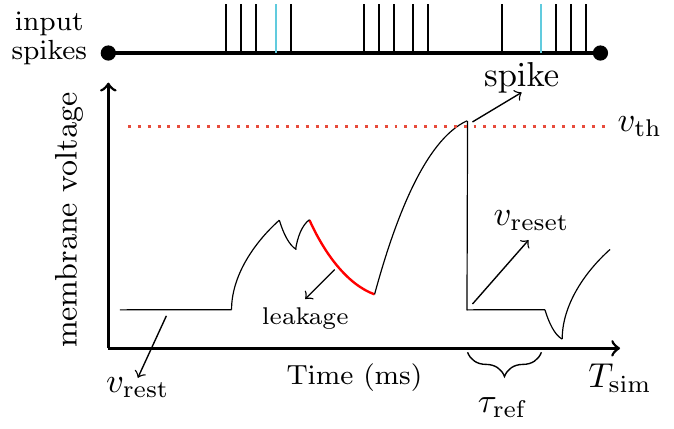}
	\caption{Given a set of input spikes (where black lines represent excitatory spikes and blue ones represent inhibitory spikes) we show how the LIF neuron parameters affect the value of the membrane potential (voltage).}
	\label{fig:lif_params}
\end{figure}

A typical representation for modelling the internal state of the spiking neuron throughout the simulation is using a dynamical system \cite{hodgkin1952quantitative,stein1965theoretical, abbott1999lapicque,izhikevich2003simple}. The most commonly used neuron model in practice is the conductance-based LIF (or Leaky Integrate and Fire) neuron \cite{abbott1999lapicque,casti2008simple}:
\begin{align}
	\tau_m \frac{d v}{d t} &= (v_{\text{rest}}-v) + g_e(E_{\text{exc}}-v) + g_i (E_{\text{inh}}-v),\label{eq:lif_conduct}\\
	\tau_{g} \frac{d g}{dt} &= -g. \label{eq:lif_conduct_syn}
\end{align}
To more realistically replicate the spiking dynamics of biological neurons, conductance-based LIF neurons split the input into an excitatory synaptic channel ($g_e$) and an inhibitory channel ($g_i$). Depending on which channel they travel through, spikes affect the sub-threshold membrane potential $v(t)$ of the neuron in different ways. We illustrate this in Fig.~\ref{fig:lif_params}, where excitatory spikes (black) increase the neuron membrane potential and inhibitory spikes (blue) decrease it.

Upon receiving an excitatory spike, the excitatory conductance $g_e$ is increased by its weight $w$. In the absence of pre-synaptic excitatory spikes, the synapse conductance decays exponentially (at a rate set by the time constant parameter $\tau_{g_e}$), as shown in (\ref{eq:lif_conduct_syn}) \cite{diehl2015unsupervised}. A similar process occurs if the pre-synaptic spike is produced by an inhibitory neuron, but where changes are enacted through the inhibitory channel parameters ($g_i$ and $\tau_{g_i}$) instead. 

Equation~\ref{eq:lif_conduct}, therefore describes the evolution of the sub-threshold membrane potential $v(t)$ over time. The \emph{leaky} characteristic of this model refers to $\tau_m$, the time constant of the neuron membrane. At each time step $\tau_m$ causes some of the current inside the neuron to \emph{leak}, decreasing the membrane potential $v(t)$ of the neuron, in the process. This is illustrated in Fig.~\ref{fig:lif_params} as leakage.

When the membrane potential $v(t)$ reaches a predefined threshold $v_{\text{th}}$ (illustrated through the dotted line in Fig.~\ref{fig:lif_params}), a spike is generated, and the membrane potential is reset to a lower value set by $v_{\text{reset}}$. Immediately after a spike, a neuron is typically unresponsive to further stimuli, meaning that no further spikes can be produced. This period is commonly known as the absolute \emph{refractory period} (the length of which is set by $\tau_{\text{ref}}$), and is defined as the minimal time difference between two spikes produced by a neuron. To ensure that the neuron's membrane stays (and begins) within realistic values $v_{\text{rest}}$ is added to the current value of $v(t)$. Finally, to set the equilibrium potential for each channel two additional membrane potentials are introduced: $E_{\text{exc}}$ (excitatory potential) and $E_{\text{inh}}$ (inhibitory potential).

As the membrane potential is reset each time its value crosses a predefined threshold, spiking neuron models are typically discontinuous \cite{paugam2012computing}, meaning that gradient descent methods cannot be used directly to train SNNs \cite{pfeiffer2018deep}. Instead, areas of discontinuity can be smoothed with estimate functions (enabling the use of gradient descent methods), or models can be trained using methods based on Hebbian learning \cite{diehl2015unsupervised,hebb1949organization,markram1997regulation,masquelier2009competitive}.

\subsection*{Predicting with spiking neural networks}
Predictions made by SNNs are uninterpretable in their standard form \cite{dayan2001theoretical}, as a list of times at which a neuron has spiked. Due to this, an interpretation process is necessary to make human-understandable predictions. 

As we currently do not understand how information is encoded and processed in the brain, there is no single interpretation method for `decoding' the information contained in a spike train \cite{dayan2001theoretical,perkel1968neural}. Interpretation methods differ in terms of the quantitative statistical measure they assume is `encoding' information \cite{perkel1968neural}. Some examples of statistical measures for which many interpretation methods have been proposed are: exact spike times, mean firing rate, and the variation of the interspike interval \cite{daley2007introduction,kass2014analysis}.

In this paper we restrict our discussion to interpretation methods that assume information is encoded in the mean firing rate of spike trains (typically known as \emph{rate coding}). This group of methods is widely used in literature \cite{diehl2015fast,liu2016benchmarking,panda2017ensemblesnn,srinivasan2018spilinc} due to their compatibility with a wide range of training approaches and robustness to the inherent stochasticity of spike trains \cite{gerstner2014neuronal,perkel1968neural,shadlen1998variable}. For in-depth discussion of how predictions can be made based on output spike trains, readers are referred to Section~\ref{sec:spiking_repr}.

\subsection*{Types of spiking neural networks}
\label{subsec:spiking_ens_types}
%

\begin{figure*}[tb]
\centering
\subfloat[]{\includegraphics[width=.48\textwidth]{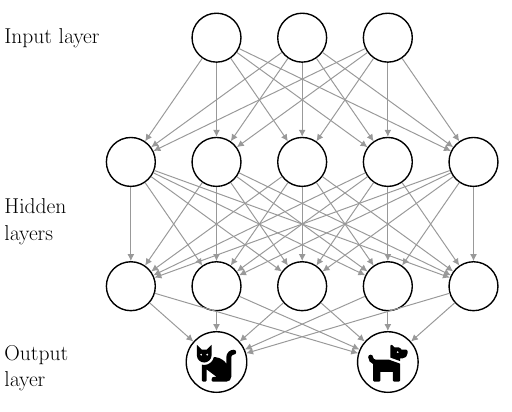}%
\label{fig:mlp_pred_arch}}
\hfil
\subfloat[]{\raisebox{0.7cm}{\includegraphics[width=.48\textwidth]{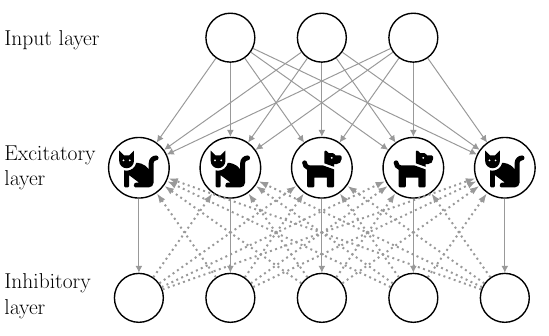}}%
\label{fig:multi_pred_arch}}
\caption{Example SNN architectures for a two-class classification task: (a) multi-layer perceptron type architecture, containing fully-connected input layer, hidden layers, and an output layer of dimensions equal to the number of possible classes (b) Multi-predictor architecture, with an input layer of 3 neurons, 5 excitatory neurons (that act as the output layer, and whose associated class is indicated through the \emph{cat} or \emph{dog} image), and 5 inhibitory neurons.}
\label{fig_sim}
\end{figure*}

While SNNs can be constructed according to traditional artificial neural network structures: CNNs, MLPs (see Fig.~\ref{fig:mlp_pred_arch}), etc., they also introduce architectures unseen in traditional neural networks. For example, because spiking neurons will not necessarily produce spikes for all inputs, some training algorithms associate multiple output neurons with the same class (such an example architecture is shown in Fig.~\ref{fig:multi_pred_arch}) \cite{bichler2012extraction,diehl2015fast,querlioz2011simulation}. Thus, before a prediction can be made, spikes produced by different output neurons (but which support the same class) have to be aggregated in some way. In the spiking literature such groups of neurons are commonly referred to as populations \cite{georgopoulos1986neuronal,hebb1949organization,sherrington1952integrative}. However, due to their benefits (collaborating to produce complex behaviors and increase certainty in predictions) such groups have also been referred to as ensembles \cite{buzsaki2010neural,deadwyler1997significance,ince2013neural,sakurai2018multiple}. 

For completeness, in Section \ref{sec:exp} we investigate both types of ensembles. Whenever we refer to combining predictions of populations of neurons we use the term ``population", and whenever we refer to combining predictions made by different SNNs we use the term ``ensemble". Populations of neurons are commonly encountered in the brain \cite{burgess1994model,georgopoulos1982relations,maunsell1983functional}. Due to this, several methods have been proposed for combining and interpreting spike trains produced by populations of spiking neurons \cite{georgopoulos1986neuronal,georgopoulos1988primate,salinas1994vector,zemel1998probabilistic}. In Section~\ref{subsec:spiking_repr_bckgr} we describe two such methods, which are used later on to evaluate our novel methods against.

As ensemble systems and populations serve the same function (collaborating to produce better predictions than an individual could) we investigate whether they would also benefit from using the same combination methods. To this end, we propose a novel interpretation method (in Section \ref{subsec:cfr}) that combines predictions of spiking neurons using a combination method that satisfies (\ref{eq:cm_req}) (guaranteeing an improvement in performance over the average single neuron). In Section \ref{sec:exp} we verify whether this guarantee can produce any benefits over existing population coding interpretation methods. 


\subsection{Training methods \& datasets} \label{subsec:training_and_datasets}
The goal of this paper is to study the performance of SNN ensembles, rather than push the state of the art for a single SNN. Therefore existing learning methods from the literature are used to train individual SNNs within an ensemble. Two methods are explored: a bio-inspired unsupervised STDP configuration (see Section~\ref{sec:d_and_c_STDP}); and the supervised gradient-based Spike Layer Reassignment in Time method (SLAYER, see Section~\ref{sec:SLAYER}). Together, these methods represent common approaches to training SNNs in the theoretical neuroscience community, meaning demonstration of the power of ensembling in conjunction with these techniques should appeal to a wide audience. Published results using these methods also exist for a range of datasets, enabling direct comparison of ensembled results against state of the art performance. 

\subsection*{Datasets}

Ensemble experiments are performed on a range of datasets to explore performance on different tasks. The MNIST dataset (images of handwritten digits) \cite{lecun-mnisthandwrittendigit-2010}, is selected for its popularity in the machine learning community, and the existence of state-of-the-art results (for both single models and ensembles of SNNs) against which to compare findings. An interesting feature of SNNs is their potential to operate efficiently on event-based input, therefore two additional datasets are explored: NMNIST (a neuromorphic version of MNIST) \cite{orchard2015converting}, and DVS Gesture (recordings of individuals performing different actions) \cite{amir2017low}.

Both MNIST and NMNIST contain 60,000 training examples and 10,000 testing examples of images (with $28\times28$ and $34\times34$ pixels respectively) illustrating one of 10 classes. Although both datasets encode the same problem, NMNIST is considered to be more difficult as saccadic motions are introduced by the sensor moving over each example during the conversion process (into spike trains). In contrast to this, the NMNIST dataset was generated by moving a Dynamic Vision Sensor on a pan-tilt unit. Each image is encoded as a spike train of `on' and `off' spikes, recorded over 300 ms.

The DVS Gesture dataset \cite{amir2017low} is also generated using an event-based sensor, encoding on/off events at a resolution of $128\times128$. The dataset contains recordings of 29 individuals performing one of 11 actions (e.g.~clapping, hand waving), each over a period of $\approx 6 \, \mathrm{s}$. Each action is recorded using a DVS camera under different lighting conditions.

\subsection*{Unsupervised Training via  STDP} \label{sec:d_and_c_STDP}
Diehl and Cook \cite{diehl2015unsupervised}, proposed an SNN architecture with a bio-inspired learning mechanism for MNIST digit recognition. It harnesses the biologically observed mechanism of spike timing dependent plasticity (STDP) \cite{bi1998synaptic}, together with conductance-based neuron models, all exhibiting an exponential time dependence. A brief summary of the approach is provided here to give context to the ensembled results, however readers are referred to the original paper \cite{diehl2015unsupervised} for additional information. 

The proposed SNN architecture is similar to that in Fig.~\ref{fig:multi_pred_arch}, containing an input layer, excitatory layer, and inhibitory layer. The number of neurons in each layer can vary, however in both the original study and this work the input layer has dimensions $28\times28$ to align with the MNIST dataset. Both the excitatory and inhibitory layers contain the same number of neurons, with increased numbers of neurons leading to increased classification accuracy. Input neurons are connected with plastic synapses to the excitatory layer in an all-to-all fashion, enabling input spike trains to excite neurons in the excitatory layer. Each neuron in the excitatory layer is then connected in a one-to-one pattern with a corresponding inhibitory neuron, with static synapses weighted such that an a spike in the excitatory layer causes a corresponding spike in the inhibitory layer. Each neuron in the inhibitory layer subsequently inhibits (with static synapses) all neurons in the excitatory layer except for its stimulating neuron, providing lateral inhibition. 

To convert the MNIST dataset to spike trains we follow the same simulation procedure as Diehl et al.~\cite{diehl2015unsupervised}. Each image is represented as a set of Poisson distributed spike trains (one for each input pixel), with maximum lengths of 350 ms. Each pixel is thus considered to be proportional to the rate of the Poisson process from which the spike train is sampled. To ensure the Poisson rates vary within realistic ranges (between 0 and 63.75 Hz in this case), the maximum pixel intensity of 255 is divided by 4. 

Synapses connecting input to excitatory neurons are plastic, with weight modifications made depending on the relative timing of presynaptic and postsynaptic spikes. A trace-based implementation is used, where $x_{pre}$ and $x_{post}$ define traces capturing the recent spike history for pre- and post-synaptic neurons respectively. On arrival of a presynaptic spike at the excitatory neuron, a weight change $\Delta w$ is calculated according to (\ref{eq:stdp_update}).
\begin{equation}
    \Delta w= \eta(x_{pre} - x_{tar})(w_{max}-w)^\mu
    \label{eq:stdp_update}
\end{equation}
Weight changes therefore occur in an online fashion, with learning rate $\eta$, an upper bound $w_{max}$, and a dependence on the previous weight defined through $\mu$. The target value of the presynaptic trace at the moment of a postsynaptic spike is defined by $x_{tar}$. The higher this value is set, the lower the synaptic weight will be -- helping ensure that presynaptic neurons which rarely lead to firing of the postsynaptic neuron are disconnected over time. 

The topology and learning mechanism sets up a winner-take-all arrangement, whereby a certain excitatory neuron will emerge as the active neuron for a given input. Once training is complete, excitatory neurons are assigned classes by presenting the entire training dataset, and observing the highest output response to a particular class. As there are many more excitatory neurons than classes, multiple excitatory neurons will be associated with the same class. Testing is then performed on the entire 10000 examples, with the predicted digit determined by averaging the responses of each neuron per class, and selecting the class with the highest mean firing rate (see Fig.~\ref{fig:multi_pred_arch}). 

In the experiments of Section~\ref{sec:exp}, the unsupervised STDP algorithm is applied to the digit recognition problem, with all simulations executed using the CPU-based BRIAN simulator \cite{Stimberg2019}. The model is configured as described above, with the number of excitatory neurons set to 400, leading to 313600 trainable parameters per SNN. In the original work~\cite{diehl2015unsupervised} this configuration achieved a reported accuracy of 82.9\%, while a higher accuracy of 95.0\% was achieved using a network with 6400 excitatory neurons (5M trainable parameters), with both results obtained with models trained over 15 cycles of the training dataset. The goal of this work is to explore the benefits of using ensembles of SNNs, and therefore the smaller less-accurate network is preferred as a test vehicle, and all results presented use only a single pass through the training dataset.

\subsection*{Supervised Training via SLAYER} \label{sec:SLAYER}
A short-coming of SNNs has been a lack of the gradient-based training methods which have made deep artificial neural networks so successful. A particular challenge has been the non-differentiable spike function, preventing direct application of the back-propagation mechanism.  Recently, however, a number of works have utilised surrogate or pseudo derivatives to overcome this challenge, producing results comparable to artificial neural networks \cite{neftci2019surrogate}. While these methods achieve impressive performance in their own right, it is interesting to understand whether performance can be further enhanced through ensembling. The SLAYER \cite{shrestha2018slayer} training method is therefore applied to train individual networks within ensembles. This particular gradient-based approach is employed due to its state of the art performance, and optimised CUDA implementation enabling fast training. 

The SLAYER (Spike LAYer Error Reassignment) algorithm solves two aspects of training SNNs via gradient-based methods: the temporal error credit reassignement, redistributing the effect of transient operations in SNNs; and the non-differentiable spike function. This not only enables back-propagation of error through the network, but acknowledges the temporal nature of SNNs to back-propagate error back in time to take into account the previous state of neurons. The non-differentiable spike function is replaced in the error back-propagation process with a probability density function representing the change in state of a neuron. SNNs are trained via supervision, with each neuron in the one-hot encoded output layer prescribed a target number of spikes based on whether it represents the correct/incorrect classification. 

The SLAYER implementation uses leaky-integrate-and-fire (LIF) neurons, with current-based exponential synapses. This simplifies (\ref{eq:lif_conduct}) to remove the $(E_{e}-v)$ term, leaving the conductance to be dependent only on the presynaptic spike train (no dependence with postsynaptic neuron potential). Weights are not split according to excitatory/inhibitory receptors, meaning all spikes are processed through the excitatory route of (\ref{eq:lif_conduct}), but with the potential for synaptic weights to become negative during training. All simulations are performed using the PyTorch SLAYER implementation \cite{shrestha2018slayer}, with the algorithm applied to two supervised learning problems: digit classification on the NMNIST dataset; and gesture classification on the DVS Gesture dataset.

For the NMNIST experiments, a four-layer fully-connected SNN is employed, containing an input layer, two hidden layers and an output layer. The number of neurons in the input and output layers are fixed at $34\times34\times2$ and 10, to match the input dimensions and number of classes respectively. The same class target as described in \cite{shrestha2018slayer} was encoded for our models (i.e.~the class with the highest spike count assigns the class, with each false class neuron assigned a target of 10 spikes, while the true class neuron is expected to spike 60 times). The original work used two hidden layers containing 500 neurons, giving a total of $1.4 \, \mathrm{M}$ trainable parameters in the network, which achieved a prediction accuracy of $98.89 \pm 0.06\%$. The goal of this work is to demonstrate the benefit of combining predictions from multiple SNNs within an ensemble, therefore the model used in this work is a scaled down version of the original model, containing the same input/output dimensions, but with two hidden layers containing only 50 neurons. This results in 118,600 trainable parameters for the network, helping speed up training time and reduce memory footprint.   


For experiments on the DVS Gesture dataset, an eight-layer convolutional SNN is employed, containing multiple convolutional and pooling layers replicating the 128x128x2-4a-16c5d-2a-32c3d-2a-512d-11 architecture used in the original SLAYER presentation \cite{shrestha2018slayer}. The input layer matches the dimensions of the DVS array giving $128\times128\times2$ inputs, each with on/off signals, enabling direct feeding of the DVS data into the neural network. A series of pooling (denoted by `a') and convolutional (denoted by `c') layers are used to process this input, before moving to a fully-connected layer (denoted by `d') and finally to the output, giving a total of 1,059,616 trainable parameters in the spiking convolutional network (SCNN). 
In the original work, this network was trained using a target output encoding specifying: false class neurons to spike 30 times; and true class neurons to spike 180 times. The DVS Gesture dataset was split, using recordings of the first 23 subjects as training data, and recordings of the last 6 subjects as test data. In both training and testing, only the first $1.5 \, \mathrm{s}$ out of $\approx6 \, \mathrm{s}$ of each sample was used, and for speed, simulation timesteps are set at $\Delta t = 5\, \mathrm{ms}$. Despite these compromises for speed, the SLAYER SCNN achieved $93.63 \pm 0.49\%$ accuracy.

As discussed for the other classification problems, the goal of this work is not to advance prediction accuracy of a single network, but instead to explore the benefits of ensembling. The SCNN described above is therefore further simplified, reducing the total trainable parameters to 54,200. While this reduction compromises performance of a single network, it greatly reduces the training time of the overall ensemble. To enable direct comparisons with the original work\cite{shrestha2018slayer}, we use the same input/output encoding, and train/test dataset split. 


\section{Interpreting individual spiking neuron predictions}
\label{sec:spiking_repr}
Before introducing strategies for combining spiking predictions within an ensemble, we first need to establish how spike trains will be interpreted (or ``decoded''). In this study we restrict the set of possible interpretation methods to those that assume the number of spikes produced in an interval encodes most of the information necessary for making predictions (i.e.~rate coding methods). We first describe existing methods commonly used for this task and then introduce our novel methods that address the concerns we raise about the former.


\subsection{Background}
\label{subsec:spiking_repr_bckgr}
Given a trained SNN (whose output layer size is equal to the number of classes), at inference time $\nclasses$ spike trains $\boldsymbol{s}=(s_1, \dots, s_{\nclasses})$ are recorded. These spike trains encode the support shown for each class. 
Most rate coding methods interpret a spike train $s_{\classind}$ as a firing rate $r_{\classind}$ (in the most general case this becomes a collection of firing rates $\boldsymbol{r}_{\classind}=(r_1,\dots,r_{\nwindows})$, if the rate is computed over multiple windows) \cite{dayan2001theoretical,gerstner2014neuronal,perkel1968neural}. Based on these firing rates (that assume a single window is used) $\boldsymbol{r}=(r_1, \dots, r_{\classind})$ a prediction is made. While such firing rates are more interpretable than the spike trains they represent, they still require some knowledge of the assumptions imposed during training. For example, should we predict the class with the highest recorded firing rate (a widely used method \cite{diehl2015fast,liu2016benchmarking,panda2017ensemblesnn,shim2016unsupervised,wu2019direct}) or the class for which the recorded firing rate is closest to some expected value (imposed by certain training approaches \cite{bohte2000spikeprop,mostafa2017supervised}, including SLAYER)? It is important to note that this limitation is shared by many time, rank order, and population codes as well \cite{dayan2001theoretical}. 

In addition to this interpretability issue, in Section \ref{sec:ensemble_bck} we show that for a combination mechanism to produce an ensemble with a guaranteed performance improvement over the average single model, the target domain has to be represented using a distribution with an exponential family. Traditionally in the machine learning literature, the target domain for classification problems is represented as a categorical distribution. However, firing rates can convey characteristics about the behavior of the spiking neurons using biologically meaningful values (unlike probability estimates) and should therefore be considered as well. In Section \ref{subsec:our_interpr_mtds} we will show how both of these representations can be produced from spike trains.

A definition for ensemble systems not encountered in traditional machine learning models is having to combine the predictions of a set of output neurons, trained in the same network, which spike for examples of the same class (an example of such an architecture is used by the Diehl and Cook models \cite{diehl2015unsupervised} and is illustrated in Fig.~\ref{fig:multi_pred_arch}). In the SNNs literature this is known as a \emph{population} \cite{anderson1994neurobiological,georgopoulos1986neuronal,hebb1949organization}, but the term of ensemble has been associated with it as well \cite{buzsaki2010neural,deadwyler1997significance,ince2013neural,sakurai2018multiple}. For completeness we compare both our novel interpretation methods (Section~\ref{subsec:our_interpr_mtds}) and our ensemble combination methods against two well known population decoding methods (the Bayes decoder and the Population vector method). Although population coding methods share more similarities with ensembles than they do with methods like HMFR, they can be applied to standard architectures (i.e.~MLPs or CNNs) as well.

Furthermore, we investigate whether combining predictions of neuron populations using an ensemble combination method that satisfies (\ref{eq:cm_req}) will lead to higher performances than those achieved by existing population coding methods. This could indicate that populations benefit from adopting the same assumptions as ensemble systems when combining predictions. For this, we introduce a novel interpretation method referred to as CFR (combined firing rates). Although we expect this method to reach its full potential when used to interpret and combine the spike trains of a population of neurons, we introduce the necessary background and notation for the CFR method to be applied on single neuron predictions in Section~\ref{subsec:poiss_glm}.

\subsection*{Highest mean firing rate (HMFR)}
Rate coding methods, and in particular, the method described below, which we call the highest mean firing rate (or HMFR) and is defined as: 
\begin{equation}
	r = \frac{1}{\ntrials} \sum_{\trialind=1}^{\ntrials} n_{\trialind}(t, t+\Delta t),
	\label{eq:hmfr}
\end{equation}
%
has produced many state-of-the-art results \cite{liu2016benchmarking,rathi2018stdp,shim2016unsupervised,srinivasan2018spilinc}. HMFR estimates the mean firing rate ($r$) produced by a neuron as an average over the spike counts $n_{\trialind}$ recorded for the same neuron, over repeated simulations (indexed using $\trialind=\{1,\dots, \ntrials\}$) of the same example. In many studies \cite{dayan2001theoretical,panzeri2001unified,diehl2015unsupervised} the firing rate is computed over small time windows of the simulation interval $T_{\text{sim}}$, to capture some temporal characteristics of the spiking pattern. While the windows are still considered independent, they are typically not large enough that the firing rate varies too quickly. Otherwise, the coding mechanism falls within temporal coding instead \cite{panzeri2001unified,gerstner2014neuronal}. Once the mean firing rate has been estimated for all output neurons, the class of the neuron with the highest firing rate is predicted.

\subsection*{Bayes decoder} 
The Bayes decoder \cite{dayan2001theoretical} was introduced as a population decoding method, but can be applied to interpret the predictions of single neurons, not part of a population as well (i.e.~output neurons in a standard MLP architecture). This method models each spiking neuron in the population as a random variable that produces spikes according to some underlying Poisson point process. This assumption is then used to estimate the conditional firing rate probability density for each neuron $j$ for each class $c$: 
\begin{equation}
    p(r_j|c) = \frac{(f_j(c)T)^{r_jT}}{(r_jT)!} \exp{(-f_j(c)T)}
\end{equation}
(where the parameters of the underlying Poisson distribution are estimated at train time from observed spike trains). To combine the individual beliefs, each neuron is assumed to be independent and the population response for each class is computed as a product over the $m$ individuals in the population: $p(r_{pop}|c) = \prod_{j=1}^m p(r_j|c)$. Bayes theorem is then used to estimate class probabilities. 

\subsection*{Population vector (PV)} 
The PV method is another population coding method, that assumes that each neuron in a population will produce a significantly higher spike count for examples of a particular class. Given spike trains produced by each neuron for examples of each class, the preference of each neuron towards each class is modeled through a set of weights (which are indicative of the expected number of spikes a neuron will produce for each class). At test time the population response for each class (i.e. $r_{pop}(c)$) is computed as a weighted average over the individual's firing rates ($r_j$) scaled by their associated class weight ($w_{jc}$): 
\begin{equation}
   r_{pop}(c) = \frac{\sum_{j=1}{m} w_{jc}r_j}{\sum_{j=1}^m r_j}.
\end{equation}
The class with the highest overall rate is then predicted.

We believe these three methods will provide a comprehensive comparison for our novel work, due to their mechanistic differences (e.g. the Bayes decoder adopts a statistical representation for the population while the PV and HMFR methods adhere to rate coding principles) and the assumptions they impose on the combination and representation of spike trains (e.g. the Bayes decoder estimates how each neuron will spike for each class, while the PV and HMFR methods assume that neurons show support for a class through high spike counts). Furthermore, all methods facilitate comparison through their use of representations that are compatible with the ensemble learning combination methods discussed in Section \ref{sec:ensemble_bck}.

\subsection{Our methods}
\label{subsec:our_interpr_mtds}

Following, we introduce two novel interpretation methods that address the interpretability issues of the HMFR and PV methods, and for the first time frame spiking neurons as generalized linear models (GLMs).

\subsection*{Representing firing rates as class probability estimates (Normalised HMFR)}
\label{subsec:rates_as_cls_prob_est}

Predicting the class of the neuron with the highest mean firing rate is equivalent to predicting the class of the neuron with the highest estimated probability, if the two methods interpret the spike train information in the same way. However, we argue that \emph{probabilities are more interpretable} because they contextualize the activity of each class with respect to the other classes. Due to this, a probability can be interpreted as the amount of support given by the model to a specific class.

To convert firing rates into class probability estimates we can use a simple normalization step. For this study we investigated three normalization methods:
\begin{itemize}
    \item Softmax ($e^{r_{\classind}} {/} \sum_{\classind=1}^{\nclasses} e^{r_{\classind}}$)
    \item Normalizing by the total activity of the output layer ($r_{\classind} {/} \sum_{\classind=1}^{\nclasses} r_{\classind}$)
    \item Normalizing using the highest recorded rate ($r_{\classind} {/} r_{max}$)
\end{itemize} 
To ensure this normalization step does not damage the model accuracy, in Section \ref{sec:exp} we evaluate the performance of SNNs (and ensembles) whose predictions have been interpreted as follows:
\begin{enumerate}
	\item Estimate the firing rate for each output neuron [using (\ref{eq:hmfr})];
	\item Combine the firing rates of populations of neurons associated with the same class (by averaging over the firing rates of neurons in each population);
	\item Normalize the class firing rates using one of the three methods above;
	\item Predict the class with the highest probability.
\end{enumerate}

The first two steps of this process are equivalent to the \emph{HMFR} (highest mean firing rate) method, while including all steps adds the normalisation, and hence defines the method termed \emph{Normalized HMFR}. Based on preliminary results \cite{neculae2020ensemble} softmax normalization is recommended in cases where the difference between the mean firing rate associated with the true class and other classes is very small, while the remaining two normalization methods should be used when high spike count fluctuations can be measured for repeated simulations of the same example. In our experiments we report the results for the normalization method that produced the highest classification accuracy.

\subsection*{Representing firing rates as Poisson means (CFR for single neurons)}
\label{subsec:poiss_glm}
In this section we introduce a novel interpretation method that models spiking neurons as GLMs. This enables us to clearly describe spiking neurons as statistical models, and facilitates the extension of this method to a population coding method (in Section~\ref{subsec:cfr}) that combines the predictions of neurons (spiking in support of the same class) using ensemble learning principles.

GLMs have been used extensively in the machine learning literature to ensure that the optimization procedure maximizes the likelihood of the model with respect to the training data for new problem domains \cite{sarle1994neural}. To our knowledge, GLMs have not been used in the same way to represent SNNs. Existing studies have employed GLMs as functional models for characterizing correlations between input stimuli and output spike trains of single (or populations of) spiking neurons \cite{pillow2008spatio,shlens2014notes,truccolo2005point}, but not of entire networks. Thus, these studies do not discuss the idea of target (or expected) firing rates.

Drawing inspiration from the traditional approach of statistically representing spike trains as samples of Poisson point processes \cite{daley2007introduction,heeger2000poisson,perkel1967neuronal}, a fitting distribution for representing the target domain of a spiking neuron is the Poisson distribution.

Given a multi-class classification problem, with $\nclasses$ classes indexed using $\classind=\{1, \dots, \nclasses\}$, we define the target domain in terms of the mean of the Poisson distribution associated with each class, as follows: 
\begin{equation}
	\lambda_{\classind} = \left\{
		\begin{array}{ll}	
			0, & y_{\exind} \neq \classind \nonumber\\
			r_{\text{max}}, & y_{\exind} = \classind. \nonumber
		\end{array}
	\right.
\end{equation}
In this case, an example $(\boldsymbol{x}_{\exind}, y_{\exind})$ whose class is $y_{\exind}=\classind$ would encode the target class as a vector of means $\boldsymbol{\lambda}=(\lambda_0,\dots,\lambda_{\nclasses})$, with all but one value equal to zero. This encoding conveys a common assumption made by rate coding interpretation methods: that large firing rates indicate strong support for a class. If a different assumption (i.e.~exact spike counts) needs to be encoded, the expected values $0$ and $r_{\text{max}}$ would be replaced with the predefined values.

Thus far we have assumed that each class is represented by a single target rate (i.e. modelling a homogeneous Poisson point process). In many cases, splitting a spike train into several windows (indexed using $\windowind$) and estimating a firing rate (represented as $r_{\classind \windowind}$) for each window can produce more accurate predictions. The target representation can easily be adapted to fit this process. Instead of having a single target rate per class we could have a vector of expected means $\boldsymbol{\lambda}_{\classind}=(\lambda_{\classind 1}, \dots, \lambda_{\classind \nwindows})$. As the time windows are non-overlapping and each spike is assumed to be independent under the Poisson point process assumption, this approach represents spike trains as samples of an inhomogeneous Poisson point process.


\section{Combining spiking neuron predictions}
\label{sec:ensemble_bck}

There are many methods for combining the predictions of a set of models \cite{polikar2006ensemble,kuncheva2014combining}. Due to the complexity of the task and the specificity of these methods, it is difficult to compare combination methods from an optimality perspective. Moreover, many studies \cite{battiti1994democracy,kittler1998combining,alkoot1999experimental,kuncheva2002theoretical} have shown the \textit{no free lunch} theorem also holds for combination methods. Thus, before choosing a method to combine the predictions of ensemble members it is important to determine what characteristics are important for the ensemble to have.  

As our aim is to improve the accuracy of SNNs, it is important that the chosen combination method produces ensembles that are guaranteed to outperform the average single model. This is a reasonable requirement in general, as ensembles are more expensive to train than single models. From the ensemble learning literature we know that a combination method will produce an ensemble with the desired performance improvement if the ensemble error can be decomposed into the average divergence of the models, and their average disagreement (commonly referred to as the Ambiguity Decomposition or AD) \cite{krogh1995neural}. 

For such a combination method to exist, specific requirements regarding the target domain and representation of predictions need to be met \cite{heskes1998selecting,webb2019joint}: 

\begin{enumerate}
	\item The target domain is represented using the same type of distribution with an exponential family for each model in the ensemble;
	\item The model parameters maximize likelihood.
\end{enumerate}

Once these conditions are met, the combination method that offers this performance guarantee (proposed in \cite{heskes1998selecting}) is the solution to:
\begin{align}
	\bar{q} &\equiv \underset{q(y|\x)}{\text{argmin}} \sum_{\modelind=1}^{\nmodels} w_{\modelind} \kl{q}{q_{\modelind}},\\
	\intertext{with the solution}
	\bar{q}(y|\x) &= \frac{1}{Z}\prod_{\modelind=1}^{\nmodels} q_{\modelind}(y|\x)^{w_{\modelind}}
	\label{eq:cm_req}
\end{align}
where $w_{\modelind}$ is the weight of each model's predictions during the combination process and $\text{KL}(q||q_{\modelind})$ measures the Kullback-Leibler divergence between the ensemble distribution and the estimate learnt by a model with index $\modelind$ in the ensemble. In other words an ensemble with minimum average `distance' (measured as KL divergence) from the models being combined is guaranteed to make more accurate predictions than the average single model. As all models estimate the same type of target distribution the ensemble distribution is guaranteed to also be the same type of distribution \cite{heskes1998selecting,webb2019joint}. This facilitates comparing the divergence between the ensemble and the target distribution, against the divergence between each model and the target distribution, allowing the AD to be formalized.

Next, we show how these theoretical principles have been implemented in two widely used combination methods that satisfy these requirements, formalizing their notation, and including their AD formulations. Then, in Section~\ref{subsec:comb_pred} we extend these principles by introducing a novel combination method for predictions that have been modelled as Poisson distributions and derive for the first time its associated AD.

\subsection{Background: Combining real-valued scores}

Given a dataset $\mathcal{D}=\{(\x_i, y_i)\}_{i=1}^N$ sampled i.i.d.~from a joint distribution $P(\boldsymbol{X},Y)$ that describes a regression problem, it is assumed that the associated conditional distribution $P(Y|\boldsymbol{X})$ is a Gaussian distribution with fixed variance, such that $y_i \sim \mathcal{N}(\mu(\x), \sigma^2)$.

When combining $\nmodels$ models trained on this dataset, where each model's estimate of the unknown distribution mean $\mu(\x)$ is represented as $f_{\modelind}$, the mean estimate for the ensemble model that satisfies (\ref{eq:cm_req}) is:
\begin{equation}
	\bar{f}(\x_i) = \frac{1}{\nmodels} \sum_{\modelind=1}^{\nmodels} f_{\modelind}(\x_i).
\end{equation}
Each predictor is weighted uniformly, though different model weights ($w_{\modelind}$) can be used such that $\sum_{\modelind=1}^{\nmodels} w_{\modelind} = 1$.

\subsection*{Ambiguity decomposition for real-valued scores}

Krogh and Vedelsby \cite{krogh1995neural} showed that for regression ensembles, where the problem is assumed to have a Gaussian distribution over its labels, at each data point the squared error of the ensemble predictor is guaranteed to be smaller than or equal to the average squared error of the individual estimators \cite{brown2004diversity}:
\begin{align}
	\label{eq:ad_regr}
	(y_{\exind}-\bar{f}(\x_{\exind}))^2 &= \frac{1}{\nmodels} \sum_{\modelind=1}^{\nmodels} (y_{\exind}-f_{\modelind}(\x_{\exind}))^2\\
	&- \frac{1}{\nmodels} \sum_{\modelind=1}^{\nmodels} (\bar{f}(\x_{\exind}) - f_{\modelind}(\x_{\exind}))^2, \nonumber
\end{align}
where the guaranteed improvement is proportional to the amount of disagreement between the predictions of the individuals, measured by the second term of (\ref{eq:ad_regr}).

\subsection{Background: Combining class probabilities}
\label{subsec:ensemble_bck_probs}
If a given dataset $\mathcal{D}=\{(\x_i,y_i)\}_{i=1}^N$ describes a classification problem, it is likely that the model predictions are represented as class probability estimates. Then, a suitable distribution to represent the target domain is the categorical distribution \cite{heskes1998selecting,miller1999critic,kuncheva2014combining}. As the categorical distribution is completely defined by the probability associated with each outcome $y \in \{1, \dots, \nclasses\}$, class labels are commonly represented as vectors of probabilities $\boldsymbol{p}=(p_1, \dots, p_{\nclasses})$ which encode the likelihood of each class such that $y_{\exind} \sim \mathcal{C}(\boldsymbol{p}(\x))$.

In this case, (\ref{eq:cm_req}) is directly applicable. Thus, given a set of $\nmodels$ models, whose estimated class probability vectors are represented as $\boldsymbol{q}_{\modelind} = (q_1, \dots, q_{\nclasses})$, the ensemble class probabilities are:  
\begin{align}
	\boldsymbol{\bar{q}} = Z^{-1} \displaystyle \prod_{\modelind=1}^{\nmodels} \boldsymbol{q}_{\modelind}^{~\frac{1}{\nmodels}}~~\text{where}~~ Z = \sum_{\classind=1}^{\nclasses} \displaystyle \prod_{\modelind=1}^{\nmodels} \boldsymbol{q}_{\modelind}^{~\frac{1}{\nmodels}},
	\label{eq:probs_cm}
\end{align}
where $Z^{-1}$ ensures that $\sum_{\classind=1}^{\nclasses} \bar{q}_{\classind} = 1$. We refer to this combination method as the normalized geometric mean (or NGM).

\subsection*{Ambiguity decomposition for class probabilities}

In this scenario, \cite{heskes1998selecting} showed that the AD takes the form:
\begin{align}
	\text{KL}(\boldsymbol{p} || \boldsymbol{\bar{q}}) &= \sum_{\modelind=1}^{\nmodels} w_{\modelind} ~\text{KL}(\boldsymbol{p}||\boldsymbol{q}_{\modelind}) - \sum_{\modelind=1}^{\nmodels} w_{\modelind} ~\text{KL}(\boldsymbol{\bar{q}}||\boldsymbol{q}_{\modelind}).
\label{eq:ad-class-steps}
\end{align}
This means that as long as the models being combined do not make identical predictions (i.e.~the second term of the AD, measuring the average disagreement from the ensemble prediction is greater than zero), the ensemble error is guaranteed to be smaller than the average error of the models being combined [i.e.~the first term in (\ref{eq:ad-class-steps})].


\subsection{Our method: Combining firing rate predictions}
\label{subsec:comb_pred}

Section~\ref{sec:spiking_repr} introduces two novel methods of representing the predictions of SNNs as samples of distributions with exponential families, satisfying the first requirement for deriving an ensemble combination method that guarantees the ensemble predictor will have less error than the average single model. The second requirement is satisfied by training the models with a method that maximizes likelihood (which is true for the two methods we investigate -- see Section~\ref{sec:spiking_neurons_bck}). In the following section we derive for the first time the combination method that offers this guaranteed improvement in accuracy and its associated AD for Poisson distributed predictions.

When predictions are represented as class probabilities we know (from Section \ref{subsec:ensemble_bck_probs}) that combining using the normalized geometric mean guarantees the ensemble will outperform the average single model. If spiking predictions are represented as firing rates instead, no such combination method has been defined in literature. Here we present a combination method that guarantees the same improvement in performance (given that firing rates are interpreted as Poisson means). For simplicity we assume that a single firing rate is estimated for each spike train (i.e.~a single window is used), but the method can easily be generalised to multiple windows (as done in our experiments).




The conditional distribution $P(Y|\boldsymbol{X})$ is represented as a set of Poisson distributions, one for each outcome $y_{\exind} \in \{1, \dots, \nclasses\}$, defined in terms of their means $\boldsymbol{\lambda}=(\lambda_1, \dots, \lambda_{\nclasses})$. These rates $\boldsymbol{\lambda}$ encode the expected activity for each class, such that $y_{\exind} \sim \mathcal{P}(\boldsymbol{\lambda}(\boldsymbol{x}))$. The ensemble rate for a class $\classind$ can therefore be derived from (\ref{eq:cm_req}) as:
\begin{equation}
	\bar{\boldsymbol{\lambda}} = \prod_{\modelind=1}^{\nmodels} \left(\hat{\boldsymbol{\lambda}}_{\modelind} \right)^{\frac{1}{\nmodels}}.
	\label{eq:rates_cm}
\end{equation}
To see how these rates $\hat{\boldsymbol{\lambda}}_{\modelind}$ are estimated from the spike trains produced by a model refer to (\ref{eq:est_poiss_mean}).

\subsection*{Ambiguity decomposition for firing rate predictions}
\label{subsec:div_ms}

The AD (\ref{eq:ad-class-steps}) for class probability predictions states that the divergence of the ensemble predictor from the target distribution $\text{KL}(\boldsymbol{p} || \bar{\boldsymbol{q}})$ is guaranteed to be smaller than the average divergence of the models from the target distribution $\sum_{\modelind=1}^{\nmodels} w_{\modelind} ~\text{KL}(\boldsymbol{p}||\boldsymbol{q}_{\modelind})$, by the average divergence of the models from the ensemble predictor $\sum_{\modelind=1}^{\nmodels} w_{\modelind} ~\text{KL}(\bar{\boldsymbol{q}}||\boldsymbol{q}_{\modelind})$. Thus, as long as the models make different mistakes, the ensemble is guaranteed to produce a performance improvement over the average single model (i.e.~the ambiguity term is greater than zero).



In the case when the target domain is represented using a set of Poisson distribution means, no AD has been derived. Following the process described in \cite{webb2019joint}, we derive the AD for this scenario as:
\begin{align}
	\text{KL}(\boldsymbol{p}(\boldsymbol{\lambda})||\boldsymbol{q}(\bar{\boldsymbol{\lambda}})) &= \frac{1}{\nmodels} \sum_{\modelind=1}^{\nmodels} (\boldsymbol{\lambda}~\log\left(\frac{\boldsymbol{\lambda}}{\boldsymbol{\hat{\lambda}}_{\modelind}}\right) + \boldsymbol{\hat{\lambda}}_{\modelind} - \boldsymbol{\lambda}) \\ 
	&- \frac{1}{\nmodels} \sum_{\modelind=1}^{\nmodels} (\boldsymbol{\bar{\lambda}}~\log\left(\frac{\boldsymbol{\bar{\lambda}}}{\boldsymbol{\hat{\lambda}}_{\modelind}}\right) +\boldsymbol{\hat{\lambda}}_{\modelind} - \boldsymbol{\bar{\lambda}}) \nonumber,
\end{align}
where $\boldsymbol{\bar{\lambda}}$ is the ensemble prediction, computed using (\ref{eq:rates_cm}), and the two terms correspond to the average model error and the ensemble diversity.

If spike trains are split into more than one window a subscript is added to each mean and the $\text{KL}(\boldsymbol{p}(\boldsymbol{\lambda}_{\windowind})||\boldsymbol{q}(\bar{\boldsymbol{\lambda}}_{\windowind}))$ is summed over each window $\windowind$.


\subsection{Alternative combination methods}
\label{subsec:alt_comb_methods}
While we have provided arguments recommending that model predictions are ensembled using combination methods guaranteeing improvements in accuracy over the average single model, no combination method can be considered optimal in all situations \cite{battiti1994democracy,alkoot1999experimental,kittler1998combining}. 
We therefore compare the presented methods against three combination methods (that do not follow the requirements defined at the start of Section~\ref{sec:ensemble_bck} and do not share their properties) commonly used in the machine learning community \cite{polikar2006ensemble,kuncheva2014combining} and in existing spiking ensemble studies \cite{shim2016unsupervised,panda2017ensemblesnn,srinivasan2018spilinc,rathi2018stdp}:

\begin{itemize}
	\item \emph{Arithmetic mean only} (AM):\\$\hat{y}_{ens} = \underset{c}{\text{argmax}} (\frac{1}{W} \frac{1}{M} \sum_{w=1}^W \sum_{m=1}^M r_{mcw})$
	\item \emph{Majority voting} (MV):\\$\hat{y}_{ens} = \underset{c}{\text{argmax}}(\underset{w}{\text{Vote}}(\underset{m}{\text{Vote}}(\{0,1\}^{c})))$
	\item \emph{Arithmetic mean \& voting} (AM\&MV):\\$\hat{y}_{ens} = \underset{c}{\text{argmax}}(\underset{w}{\text{Vote}}(\underset{c}{\text{argmax}}(\frac{1}{M}\sum_{m=1}^M r_{mcw})))$
\end{itemize}

We included the AM\&MV combination method (which is very similar to the other two: MV and AM only) because we expected that the way in which predictions are aggregated over independent windows leads to significant differences in the final prediction.


\subsection{Our method: CFR for populations}
\label{subsec:cfr}

In this section we return to the topic of interpreting spiking neuron predictions. We propose a novel population coding method that adopts the ensemble principles described in this section with the aim of investigating the similarity between populations and ensembles. This method adopts the assumptions and target representation introduced in Section~\ref{subsec:poiss_glm} (specifically CFR for single neurons), idealizing output spike trains as inhomogeneous Poisson point processes. Building on the notation and assumptions introduced there, we generalize the CFR method for single neurons to interpret spike trains produced by populations of neurons. The difference between this method and existing population coding methods (like the Bayes decoder and PV) is that neurons are represented as GLM models, which enables us to combine predictions of neurons in a population using the ensemble principles discussed above. Thus, in Section~\ref{subsec:ens_pop_exp} we explore whether populations serve the same functions as ensembles, and therefore benefit from being constructed in the same way (e.g.~can diversity in a population improve the accuracy of the population?).  

Given many more output neurons than classes, where $|\npred_{\classind}|$ records the number of output neurons spiking in support of class $\classind$, we estimate the mean of the Poisson process underlying the spiking behavior of an output neuron $\predind$ as:
\begin{equation}
	\hat{\lambda}_{\predind \windowind} = \frac{1}{\ntrials} \sum_{\trialind=1}^{\ntrials} n_{\predind \windowind \trialind},
	\label{eq:est_poiss_mean}
\end{equation}
averaging over the spike counts $n_{\predind \windowind \trialind}$ recorded for neuron $\predind$ during the simulation window $\windowind$ [which represents the time period $(t_{\windowind}, t_{\windowind}+\Delta t)$], during the $\trialind$-th simulation of the same example $(\boldsymbol{x}_{\exind}, y_{\exind})$. Multiple simulations are necessary to minimize the effect of fluctuations observed in spike trains on the firing rates estimated.
 
To estimate a single Poisson mean per class, these rates are combined using the combination method derived in (\ref{eq:rates_cm}), which in this notation becomes:
\begin{equation}
	\bar{\lambda}_{\windowind} = \prod_{\predind \in \npred_{\classind}} {\hat{\lambda}_{\predind \windowind}}^{\frac{1}{|\npred_{\classind}|}},
\end{equation}
as it produces a better estimate of the expected population Poisson process mean than the average single neuron estimate. At the end of this process, we have estimated a vector of Poisson means for each class: $\boldsymbol{\bar{\lambda}}_{\classind} = (\bar{\lambda}_{\classind 1}, \dots, \bar{\lambda}_{\classind \nwindows})$. 

Depending on the assumption encoded in the target rates (some examples of which are discussed in Section \ref{sec:spiking_repr}) the prediction process might vary. In the most general case, if output neuron populations are expected to spike with a predefined frequency $\boldsymbol{\lambda} = (\boldsymbol{\lambda}_{1}, \dots, \boldsymbol{\lambda}_{\nwindows})$, the class with the smallest divergence ($\text{KL}(\boldsymbol{p}(\boldsymbol{\lambda})||\boldsymbol{q}(\boldsymbol{\bar{\lambda}}))$) from the expectation would be predicted. 

In the following experiments we adopt the simpler and most widely used assumption in practice, that the highest spiking frequency indicates the strongest class support (also adopted by the HMFR method). Thus, in each window the class with the highest firing rate is predicted, such that $\hat{y}_{\windowind} = \underset{\classind}{\text{argmax}}(\bar{\lambda}_{1 \windowind}, \dots, \bar{\lambda}_{\nclasses \windowind})$. Then, the class with the most votes across all windows can be predicted (using a majority vote).

\section{Experiments}
\label{sec:exp}

The experimental aims investigated in the following sections demonstrate: the effect of the choice of interpretation method on the model accuracy (evaluated in Section \ref{subsec:interpr_choice_exp}), the importance of the combination mechanism and its properties when combining predictions of spiking networks (verified in Section \ref{subsec:comb_pred_exp}), and our expectation that ensemble systems and populations of neurons benefit from making different assumptions (discussed in Section \ref{subsec:ens_pop_exp}).

\subsection{Effect of spike-train interpretation method}
\label{subsec:interpr_choice_exp}

The choice of spike-train interpretation method is explored by training single SNN models and processing the output according to the methods described in Section~\ref{sec:spiking_repr}. All reported results represent the average over 5 independent runs, obtained using the best window size and normalization method (for Normalized HMFR) \cite{neculae2020ensemble}, and with synaptic weights randomly initialized via unique seeds and trained independently according to the methods described in Section~\ref{subsec:training_and_datasets}. Each of the spike train interpretation methods of Section~\ref{sec:spiking_repr} is applied to an individual model, with results presented in Table~\ref{table:interpr_mtd_effect} alongside baseline results from the literature for comparison. 

\begin{table*}[!t]
	\renewcommand{\arraystretch}{1.3}
	\caption{Evaluating the effect of spike train interpretation methods.}
	\label{table:interpr_mtd_effect}
	\centering
	\begin{tabular}{@{} l l l l l @{}}
		\toprule
		Dataset & Method & Parameters & Interpret & Accuracy\\
		\midrule
		\multirow{6}{*}{MNIST} & \multirow{6}{*}{USTDP \cite{diehl2015unsupervised}} & 5\thinspace017\thinspace600 (original) & HMFR & 95.00\%\\
		\cmidrule{3-5}
		 &  & \multirow{5}{*}{\shortstack{\textbf{313\thinspace600 (ours)}}} & HMFR & $ 67.63 \pm 0.11\%$ \\
		 &  & & \textbf{Bayes decoder} & $\mathbf{93.62 \pm 0.01\%}$\\
		 &  & & Norm.~HMFR & $ 67.63 \pm 0.11\%$\\
		 &  & & CFR & $ 68.51 \pm 0.11\%$\\
		 &  & & PV & $ 66.59 \pm 0.11\%$\\
		\midrule
		\multirow{6}{*}{NMNIST} & \multirow{6}{*}{SLAYER \cite{shrestha2018slayer}} & 1\thinspace411\thinspace000 (original SMLP) & HMFR & $98.89 \pm 0.06\%$\\
		\cmidrule{3-5}
        & & \multirow{5}{*}{\shortstack{\textbf{118\thinspace600 (ours)}}} & HMFR & $87.48 \pm 0.02\%$\\
        & & & \textbf{Bayes decoder} & $\mathbf{99.43 \pm 0.00\%}$\\
        & & & Norm.~HMFR & $87.56 \pm 0.02\%$\\
        & & & CFR & $87.48 \pm 0.02\%$\\
        & & & PV & $87.48 \pm 0.02\%$\\
		\midrule
		\multirow{6}{*}{\parbox{0.9cm}{DVS Gesture}} & \multirow{6}{*}{SLAYER \cite{shrestha2018slayer}} & 1\thinspace059\thinspace616 (original SCNN) & HMFR & $93.64 \pm 0.49\%$\\ 
		\cmidrule{3-5}
        &  & \multirow{5}{*}{\shortstack{\textbf{54\thinspace200 (ours)}}} & HMFR & $68.12 \pm 0.04\%$\\
        &  &  & \textbf{Bayes decoder} & $\mathbf{97.49 \pm 0.01\%}$\\
        &  &  & Norm.~HMFR & $68.12 \pm 0.04\%$\\
        &  &  & CFR & $68.12 \pm 0.04\%$\\
        &  &  & PV & $65.52 \pm 0.03\%$\\
		\bottomrule
	\end{tabular}
\end{table*}

To show the full advantage of ensemble systems, the evaluated models are intentionally kept small to reduce the number of trainable parameters, and by extension their training times. Thus, it should not be expected that a single model evaluated with the same interpretation method as in the original paper will be able to outperform the accuracy of the original model. For example, the HMFR interpretation of the USTDP model trained on MNIST achieves a single model performance of $67.63\%$, whereas the original paper achieved $82.9\%$ accuracy with the same network architecture. This difference is due to results here using only a single pass through the training dataset during learning, while the original work used three passes. Despite the small models and simplified training processes, we expect certain interpretation methods (particularly the Bayes decoder) to outperform the accuracy produced with the HMFR method for our models, as different characteristics of the spike trains (e.g.~information encoded in the exact timings of spikes) are exploited. Furthermore, we expect that when these weaker models are combined, the ensembles will produce higher accuracies than the original models.

As shown in Table~\ref{table:interpr_mtd_effect}, the choice of interpretation method has a significant effect on model accuracy, with the Bayes decoder method producing the highest accuracies of all methods considered. Furthermore, for the models trained using SLAYER, the Bayes decoder has produced higher accuracies than those reported in the original paper, despite using much smaller models (or reduced training cycles in the case of USTDP) -- see Section~\ref{sec:spiking_neurons_bck}. This shows that choosing a powerful interpretation method can not only improve the accuracy of state-of-the-art models, but also reduce significantly the size of models necessary to produce such performance (reductions in trainable parameters relative to original models of $16\times$, $12\times$ and $20\times$ respectively for the tasks: MNIST, NMNIST and DVS Gesture).

To understand the differences recorded above, we need to look at the mechanisms employed by the interpretation methods. The Bayes decoder estimates the parameters of the inhomogeneous Poisson point process that fits the observed spike trains. Using this point process estimate and its underlying distribution, the support for each class is represented as a probability. This approach has the advantage that the expected spiking frequency for each output neuron is estimated from training data, which can more closely match the behavior of the trained model than the general assumption that high spiking frequencies indicate strong support (adopted by the other methods). In addition to this, the fact that the Bayes decoder shows a clear preference towards small windows (10 ms vs 350 ms preferred by the other methods \cite{neculae2020ensemble}) suggests that it can capture information encoded in the exact times of spikes (which becomes less obvious as the window size is increased). In contrast, the other 4 interpretation methods require much larger window sizes because they all rely on estimating firing rates which require more observations to produce good estimates of the spiking activity. 

\subsection{Effect of ensemble prediction combination method}
\label{subsec:comb_pred_exp}
The choice of combination method is explored by training an ensemble of SNNs, with each individual model using the Bayes decoder interpretation method, and combining the outputs according to the methods described in Section~\ref{sec:ensemble_bck}. All reported results represent the average over 5 independent runs, obtained using the best window size and normalization method (for Normalized HMFR) \cite{neculae2020ensemble}. Each ensemble combines the predictions of 5 models with the same architecture, whose weights have been initialized with a different random seed, and which have been trained independently.

So far we have given arguments to recommend that model predictions are ensembled using combination methods that guarantee improvements in accuracy over the average single model, however no combination method can be considered optimal in all situations \cite{battiti1994democracy,alkoot1999experimental,kittler1998combining}. Due to this, in Table \ref{table:ens_comb_comp} we evaluate our recommended combiners (i.e.~the normalized geometric mean, or NGM, for the Bayes decoder ensembles), against three combination methods that are commonly used in the machine learning community \cite{polikar2006ensemble,kuncheva2014combining} and in existing spiking ensemble studies \cite{shim2016unsupervised,panda2017ensemblesnn,srinivasan2018spilinc,rathi2018stdp}: Arithmetic Mean (AM), Majority Voting (MV) and Arithmetic mean \& Majority Voting (AM\&MV) -- see Section.~\ref{subsec:alt_comb_methods}. The AM\&MV combination method is included (despite its similarity to: MV and AM only), due to the way in which its predictions are aggregated over independent windows leading to significant differences in the final prediction.

\begin{table*}[!t]
	\renewcommand{\arraystretch}{1.3}
	\caption{Evaluating the effect of combination methods on overall ensemble performance.}
	\label{table:ens_comb_comp}
	\centering
	\begin{tabular}{@{} l l l l l l @{}}
		\toprule
		Dataset & Method & Parameters & Interpret & Combine & Accuracy\\
		\midrule
		\multirow{8}{*}{MNIST} & \multirow{5}{*}{USTDP \cite{diehl2015unsupervised}} & 5\thinspace017\thinspace600 (original) & HMFR & - & 95.00\%\\
		\cmidrule{3-6}
		 &  & \multirow{4}{*}{\shortstack{\textbf{1\thinspace568\thinspace000 (ours - ens)}}} & \multirow{4}{*}{\shortstack{\textbf{Bayes}\\ \textbf{decoder}}} & \textbf{NGM} & $\mathbf{98.71 \pm 0.00\%}$\\
		 &  & & & AM & $ 94.72 \pm 0.00\%$\\
		 &  & & & MV & $ 97.74 \pm 0.00\%$\\
		 &  & & & \textbf{AM\&MV} & $\mathbf{98.71 \pm 0.00\%}$\\
		 \cmidrule{3-6}
         & SCNN \cite{fang2020exploiting} & 1\thinspace675\thinspace552 (original) & Norm.~HMFR & -- & \textbf{99.46}\% \\ 
         & SNN \cite{cheng2020finite} & 397\thinspace000 (original) & HMFR & -- & $98.69 \pm 0.03\%$ \\ 
         & SpiLinC \cite{srinivasan2018spilinc} & 4\thinspace866\thinspace048 (original - ens) & HMFR & AM & 90.90\% \\ 
		 & Multimodal SNN \cite{rathi2018stdp} &  10\thinspace035\thinspace200 (original - ens) & HMFR & AM & 98.00\% \\ 
		 & EnsembleSNN \cite{panda2017ensemblesnn} & 219\thinspace520 (original - ens) & HMFR & AM & 86.98\% \\ 
		\midrule
		\multirow{5}{*}{NMNIST} & \multirow{5}{*}{SLAYER \cite{shrestha2018slayer}} & 1\thinspace411\thinspace000 (original SMLP) & HMFR & - & $98.89 \pm 0.06\%$\\
		\cmidrule{3-6}
        &  & \multirow{4}{*}{\shortstack{\textbf{593\thinspace000 (ours - ens)}}} & \multirow{4}{*}{\shortstack{\textbf{Bayes}\\ \textbf{decoder}}}        
        & \textbf{NGM} & $\mathbf{100.0 \pm 0.00\%}$\\
        &  & & & \textbf{AM} & $99.80 \pm 0.00\%$\\
        &  & & & MV & $\mathbf{100 \pm 0.00\%}$\\
		&  & & & \textbf{AM\&MV} & $\mathbf{100.0 \pm 0.00\%}$\\
		\cmidrule{3-6}
		& SNN \cite{kugele2020efficient} & 300\thinspace000 (original) & HMFR & -- & $99.56\pm 0.01\%$ \\ 
		& SCNN \cite{fang2020exploiting} & 1\thinspace394\thinspace752 (original) & Norm.~HMFR & -- & 99.39\% \\ 
		\midrule
		\multirow{5}{*}{\parbox{0.9cm}{DVS Gesture}} & \multirow{5}{*}{SLAYER \cite{shrestha2018slayer}} & 1\thinspace059\thinspace616 (original SCNN) & HMFR & - & $93.64 \pm 0.49\%$\\ 
		\cmidrule{3-6}
        &  & \multirow{4}{*}{\shortstack{\textbf{271\thinspace000 (ours - ens)}}} & \multirow{4}{*}{\shortstack{\textbf{Bayes}\\ \textbf{decoder}}} & \textbf{NGM} & $\mathbf{99.09 \pm 0.00\%}$\\
        &  & & & AM & $95.92 \pm 0.01\%$\\
        &  & & & \textbf{MV} & $\mathbf{99.09 \pm 0.00\%}$\\
		&  & & & \textbf{AM\&MV} & $\mathbf{99.09 \pm 0.00\%}$\\
		\cmidrule{3-6}
		& SNN \cite{kugele2020efficient} & 800\thinspace000 (original) & HMFR & -- & $95.68\pm 0.32\%$ \\ 
		& SCNN \cite{fang2020exploiting} & 690\thinspace816 (original) & Norm.~HMFR & -- & 96.09\% \\ 
		\bottomrule
	\end{tabular}
\end{table*}

As shown in Table \ref{table:ens_comb_comp} on NMNIST and DVS Gesture our ensembles produce state-of-the-art results, improving upon the performance of existing methods by large margins. On the MNIST dataset, our ensembles outperform all existing spiking ensemble models and single spiking networks, except for the spiking CNN (SCNN) proposed in \cite{fang2020exploiting}. However, taking into account the differences between training approaches (i.e.~USTDP is an unsupervised training routine) and the size of our individual networks (313\thinspace600 parameters) it is clear that our ensembles could produce even better results given larger networks and more training iterations. As our aim is not to reach a new state-of-the-art performance, but to demonstrate how reliably ensembles can outperform complex single models when their outputs are interpreted and combined as shown here, these small networks are better suited. 

In each case the ensembles produced large improvements in accuracy over the original models they are compared against using much smaller architectures. While in most cases all combination methods produced an improvement over the accuracy of the single model reported in Table \ref{table:interpr_mtd_effect} (with the exception of the AM combiner on the DVS Gesture dataset), the NGM (normalized geometric mean) and AM\&MV combiners have reliably produced the best accuracies on all three datasets. 

The AM\&MV combiner was designed to tackle some specific weaknesses observed in the AM and MV combiners when used separately. For example, the AM combiner is easily skewed by windows with high spike counts, which in many situations is beneficial (hence the good performance produced by these ensembles). However, for interpretation methods that benefit from splitting the simulation interval into many windows (as the Bayes decoder does) having a small subset of windows drown out the rest of the predictions is not beneficial, and a method such as MV (which assigns equal weight to each window) might be more suitable. In this case AM\&MV perfectly addresses these issues, however to gain such insights many experiments need to be run and evaluated in depth, which is not always possible or desirable. Although in the experiments presented here AM\&MV produces very good results it cannot be guaranteed that on different datasets, architectures, or training approaches the same results would be produced. Thus, we argue that a combination method that guarantees an improvement over the average single model performance (such as the NGM method) should be preferred and used over alternatives.

\begin{table}[!t]
	\renewcommand*{\arraystretch}{1.3}
	\caption{AD terms for ensembles combined using the methods in Section \ref{subsec:ensemble_bck_probs} and Section \ref{subsec:comb_pred} interpreted using all methods considered}
	\label{table:ad_ens}
	\centering
	\begin{tabular}{@{} p{0.7cm} p{1.4cm} p{1.5cm} p{1.7cm} p{2cm}  @{}}
		\toprule
		Dataset & Method & Interpret & Avg. model error & Avg. ambiguity \\
		\midrule
		\multirow{5}{*}{MNIST} & \multirow{5}{*}{USTDP \cite{diehl2015unsupervised}} & HMFR & $6.07\times 10^{-7}$ & $4.38\times 10^{-7}$ \\ 
		& & Bayes decoder & $6.88\times 10^{-1}$ & $3.15\times 10^{-1}$ \\
		& & Norm.~HMFR & $5.61\times 10^{-10}$ & $1.70\times 10^{-10}$ \\ 
		& & CFR & $1.26\times 10^{-5}$ & $8.64\times 10^{-6}$ \\ 
		& & PV & $1.08\times 10^{-7}$ & $6.79\times 10^{-8}$ \\ 
		\midrule
        \multirow{5}{*}{NMNIST} & \multirow{5}{*}{SLAYER \cite{shrestha2018slayer}} & HMFR & $5.86\times 10^{-1}$ & $5.59\times 10^{-1}$ \\ 
		& & Bayes decoder & $5.07\times 10^{-1}$ & $4.35\times 10^{-1}$ \\
		& & Norm.~HMFR & $6.05\times 10^{-1}$ & $7.62\times 10^{-2}$ \\ 
		& & CFR & $5.86\times 10^{-1}$ & $5.59\times 10^{-1}$ \\ 
		& & PV & $3.84\times 10^{-3}$ & $2.47\times 10^{-5}$ \\ 
		\midrule
        \multirow{5}{*}{\parbox{0.9cm}{DVS Gesture}} & \multirow{5}{*}{SLAYER \cite{shrestha2018slayer}} & HMFR & $1.50\times 10^{-2}$ & $1.50\times 10^{-2}$ \\ 
		& & Bayes decoder & $3.73\times 10^{-3}$ & $3.73\times 10^{-3}$ \\
		& & Norm.~HMFR & $1.43\times 10^{0}$ & $1.43\times 10^{0}$ \\ 
		& & CFR & $1.50\times 10^{-1}$ & $1.07\times 10^{-1}$ \\ 
		& & PV & $1.60\times 10^{-3}$ & $1.60\times 10^{-3}$ \\ 
		\bottomrule
	\end{tabular}
\end{table}

Furthermore, NGM is the only combination method (out of the 4 included in Table \ref{table:ens_comb_comp}) that constructs ensembles for which the average error and diversity can be measured and used to gain further insights into the ensembles constructed. From Table \ref{table:ad_ens} we see that certain interpretation methods are better at conveying differences between output spike trains in the predictions they make, leading to ensembles with higher diversity (reflected in the average ambiguity). It is important to note that comparisons are made by discussing the size of the ambiguity term relative to the size of the average model error, as direct comparisons between the ensembles are not possible.

The similar values for accuracy and diversity recorded for all ensembles combining firing rates (i.e.~HMFR, CFR, and PV), could show that the representation they adopt (firing rates) is insufficiently expressive. While this is likely true to some degree, the fact that the HMFR and Normalized HMFR single model and ensemble average accuracy are identical suggests that the interpretation method has a larger effect on the amount of diversity that is conveyed through the decoded predictions. For example, the fact that the Bayes decoder performs better when using smaller windows is possible in part due to its mechanism of estimating class probabilities and combining these probabilities. When a similar process is attempted using firing rates, a small subset of windows that contain most of the spikes recorded end up dictating the final prediction, which does not happen when using probabilities as they contextualize the activity for other classes as well (hence they are normalized by the total activity recorded).

\subsection{Should populations of neurons be treated as ensembles?}
\label{subsec:ens_pop_exp}

As the models trained with the unsupervised STDP method \cite{diehl2015unsupervised} are the only ones (out of the ones trained) making use of populations of output neurons to make predictions, we restrict our discussion here only to them. 

\begin{figure}[!t]
	\centering
	\includegraphics[width=0.5\textwidth]{{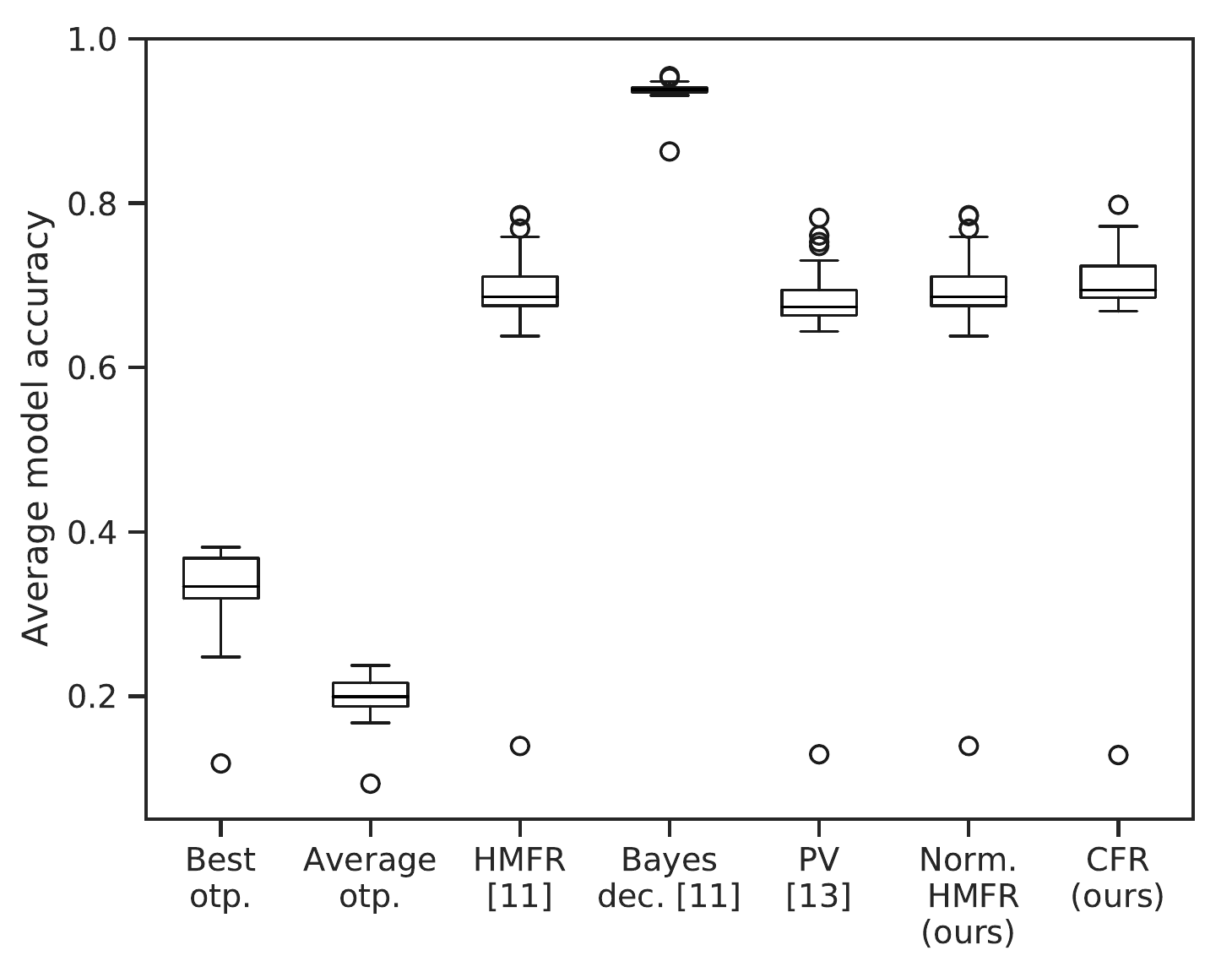}}
\caption{Combining and interpreting predictions of spiking neurons (trained on MNIST) according to ensemble learning principles leads to similar or worse predictions to those produced using population coding methods (PV or the Bayes decoder).} 
\label{fig:ens_vs_pop_assumptions}
\end{figure}

Fig.~\ref{fig:ens_vs_pop_assumptions} shows that ensemble systems and populations do not follow the same assumptions. As the task being learnt is too complex for a single neuron to solve, it is natural that the predictions made with the best (Best otp.) or average (Average otp.) single neuron of each population are outperformed by all other interpretation methods, regardless of their categorization as population coding methods (Bayes decoder, PV, and CFR), or not (HMFR and Norm.~HMFR). 

The desirable properties of the Bayes decoder method (preference for small window sizes, estimating the expected target activity from training data) discussed so far, which are not shared by other methods, address the aim of this investigation. Populations of neurons do not benefit from being treated as ensemble systems because they do not share many of the properties of the ensemble individuals. A single neuron in a population cannot be treated as a weak learner because it has not been trained to be capable of distinguishing between classes on its own, it relies on the support of the other neurons in the population to make correct predictions. From this perspective a population of neurons is more similar to a tightly correlated ensemble (e.g.~ensembles trained with modular loss \cite{webb2019joint}), than the ensembles discussed here.
\section{Conclusions}
\label{sec:discussion}



This work explores the application of ensemble learning theory to spiking neural networks (SNNs), with the goal of improving performance of these bio-inspired algorithms. Performance is assessed on common tasks from the literature, including the image-recognition standard MNIST, along with event-based image recognition and gesture datasets incorporating temporal information. Two learning methods from the literature are utilised to train individual networks within ensembles: a bio-inspired unsupervised STDP approach; and a gradient-based technique using pseudo derivatives to overcome the spike discontinuity and enable training via error back-propagation. While a great number of methods now exist for training SNNs, these techniques are representative of both the neuroscience and machine learning communities, and demonstrate the power of ensembling when applied to both domains. 

We first show that the method used to interpret spike trains into human-understandable predictions (Section~\ref{sec:spiking_repr}) has a significant effect on the accuracy of the models, with methods such as the Bayes decoder producing state-of-the-art performance with much smaller models than previously reported (reductions of trainable parameters relative to published models of $16\times$, $12\times$ and $20\times$ respectively for the tasks: MNIST, NMNIST and DVS Gesture). In addition to this, by decomposing the error of the ensembles evaluated here we show that certain interpretation methods (namely the Bayes decoder) are more capable of conveying differences in output spike trains into the predictions they make, leading to ensembles with larger diversities (and larger increases in ensemble accuracy by consequence). 

We then demonstrate how to ensemble SNNs with the guarantee that the ensemble predictor will outperform the average single model (Section \ref{sec:ensemble_bck}), through extension of the set of combination methods with this property to predictions that are Poisson distributed. Furthermore, we derive for the first time the Ambiguity Decomposition for this combination method, which expresses the average ensemble error in terms of the average model error and the average ensemble ambiguity (indicative of ensemble diversity). Comparing the ensemble combination methods advocated here (i.e.~the normalized geometric mean for class probabilities and the geometric mean combiner for firing rates), against typical combination methods used in literature (taking an arithmetic mean or majority vote over predictions), and a novel method designed to address specific issues (the arithmetic mean \& majority vote method, which is a combination of the former two), we show that our methods should be preferred in all cases. Ensemble predictions surpass state-of-the-art for all results presented, achieving classification accuracies of $98.71\%$, $100.0\%$, and $99.09\%$, on the MNIST, NMNIST and DVS Gesture datasets respectively. Furthermore, the ensemble systems produced an overall reduction in trainable parameters relative to the originally published models, exhibiting reductions of $3.2\times$, $2.4\times$ and $3.9\times$ respectively for the three tasks. 

Finally, we argue and empirically demonstrate that ensembles and populations of neurons benefit from different strategies when combining predictions/spike trains. This is because all models in an ensemble have the same definition for the target, while in a population no such definition exists. Each neuron develops an internal target expectation (i.e.~learns to show support for a class) and produces spikes according to that definition. Benchmarking our ensembles against existing literature we show that an ensemble can produce higher accuracy than the average single model (as reported in the original papers), with a reduction in computation costs (as ensembles benefit from using weaker models).

A natural continuation of this work would be to investigate how diversity can be encouraged in ensembles of SNNs \cite{neculae2020ensemble}. From implicit diversity mechanisms, to explicit methods of constructing diverse ensembles \cite{brown2004diversity}, such methods have been insufficiently explored for SNNs and have the potential to further reduce the gap between conventional and SNNs. 
\ifCLASSOPTIONcaptionsoff
  \newpage
\fi



\bibliographystyle{IEEEtran}
\bibliography{IEEEabrv,refs.bib}

\begin{thebibliography}{10}
\providecommand{\url}[1]{#1}
\csname url@samestyle\endcsname
\providecommand{\newblock}{\relax}
\providecommand{\bibinfo}[2]{#2}
\providecommand{\BIBentrySTDinterwordspacing}{\spaceskip=0pt\relax}
\providecommand{\BIBentryALTinterwordstretchfactor}{4}
\providecommand{\BIBentryALTinterwordspacing}{\spaceskip=\fontdimen2\font plus
\BIBentryALTinterwordstretchfactor\fontdimen3\font minus
  \fontdimen4\font\relax}
\providecommand{\BIBforeignlanguage}[2]{{%
\expandafter\ifx\csname l@#1\endcsname\relax
\typeout{** WARNING: IEEEtran.bst: No hyphenation pattern has been}%
\typeout{** loaded for the language `#1'. Using the pattern for}%
\typeout{** the default language instead.}%
\else
\language=\csname l@#1\endcsname
\fi
#2}}
\providecommand{\BIBdecl}{\relax}
\BIBdecl

\bibitem{he2016deep}
K.~He, X.~Zhang, S.~Ren, and J.~Sun, ``Deep residual learning for image
  recognition,'' in \emph{Proceedings of the IEEE conference on computer vision
  and pattern recognition}, 2016, pp. 770--778.

\bibitem{touvron2019fixing}
H.~Touvron, A.~Vedaldi, M.~Douze, and H.~J{\'e}gou, ``Fixing the train-test
  resolution discrepancy,'' \emph{arXiv preprint arXiv:1906.06423}, 2019.

\bibitem{averbeck2006neural}
B.~B. Averbeck, P.~E. Latham, and A.~Pouget, ``Neural correlations, population
  coding and computation,'' \emph{Nature reviews neuroscience}, vol.~7, no.~5,
  p. 358, 2006.

\bibitem{averbeck2004coding}
B.~B. Averbeck and D.~Lee, ``Coding and transmission of information by neural
  ensembles,'' \emph{Trends in neurosciences}, vol.~27, no.~4, pp. 225--230,
  2004.

\bibitem{kopell2011neuronal}
N.~Kopell, M.~A. Whittington, and M.~A. Kramer, ``Neuronal assembly dynamics in
  the beta1 frequency range permits short-term memory,'' \emph{Proceedings of
  the National Academy of Sciences}, vol. 108, no.~9, pp. 3779--3784, 2011.

\bibitem{o2006biologically}
R.~C. O'Reilly, ``Biologically based computational models of high-level
  cognition,'' \emph{science}, vol. 314, no. 5796, pp. 91--94, 2006.

\bibitem{pillow2008spatio}
J.~W. Pillow, J.~Shlens, L.~Paninski, A.~Sher, A.~M. Litke, E.~Chichilnisky,
  and E.~P. Simoncelli, ``Spatio-temporal correlations and visual signalling in
  a complete neuronal population,'' \emph{Nature}, vol. 454, no. 7207, p. 995,
  2008.

\bibitem{gerstner2002spiking}
W.~Gerstner and W.~M. Kistler, \emph{Spiking neuron models: Single neurons,
  populations, plasticity}.\hskip 1em plus 0.5em minus 0.4em\relax Cambridge
  university press, 2002.

\bibitem{hodgkin1952quantitative}
A.~L. Hodgkin and A.~F. Huxley, ``A quantitative description of membrane
  current and its application to conduction and excitation in nerve,''
  \emph{The Journal of physiology}, vol. 117, no.~4, pp. 500--544, 1952.

\bibitem{liu2014event}
S.-C. Liu, T.~Delbruck, G.~Indiveri, A.~Whatley, and R.~Douglas,
  \emph{Event-based neuromorphic systems}.\hskip 1em plus 0.5em minus
  0.4em\relax John Wiley \& Sons, 2014.

\bibitem{benosman2013event}
R.~Benosman, C.~Clercq, X.~Lagorce, S.-H. Ieng, and C.~Bartolozzi,
  ``Event-based visual flow,'' \emph{IEEE transactions on neural networks and
  learning systems}, vol.~25, no.~2, pp. 407--417, 2013.

\bibitem{mozafari2018combining}
M.~Mozafari, M.~Ganjtabesh, A.~Nowzari-Dalini, S.~J. Thorpe, and T.~Masquelier,
  ``Combining stdp and reward-modulated stdp in deep convolutional spiking
  neural networks for digit recognition,'' \emph{arXiv preprint
  arXiv:1804.00227}, 2018.

\bibitem{davies2021advancing}
M.~Davies, A.~Wild, G.~Orchard, Y.~Sandamirskaya, G.~A.~F. Guerra, P.~Joshi,
  P.~Plank, and S.~R. Risbud, ``Advancing neuromorphic computing with loihi: A
  survey of results and outlook,'' \emph{Proceedings of the IEEE}, vol. 109,
  no.~5, pp. 911--934, 2021.

\bibitem{amir2017low}
A.~Amir, B.~Taba, D.~Berg, T.~Melano, J.~McKinstry, C.~Di~Nolfo, T.~Nayak,
  A.~Andreopoulos, G.~Garreau, M.~Mendoza \emph{et~al.}, ``A low power, fully
  event-based gesture recognition system,'' in \emph{Proceedings of the IEEE
  Conference on Computer Vision and Pattern Recognition}, 2017, pp. 7243--7252.

\bibitem{Goodfellow-et-al-2016}
I.~Goodfellow, Y.~Bengio, and A.~Courville, \emph{Deep Learning}.\hskip 1em
  plus 0.5em minus 0.4em\relax MIT Press, 2016.

\bibitem{kelley1960gradient}
H.~J. Kelley, ``Gradient theory of optimal flight paths,'' \emph{Ars Journal},
  vol.~30, no.~10, pp. 947--954, 1960.

\bibitem{schmidhuber2015deep}
J.~Schmidhuber, ``Deep learning in neural networks: An overview,'' \emph{Neural
  networks}, vol.~61, pp. 85--117, 2015.

\bibitem{paugam2012computing}
H.~Paugam-Moisy and S.~Bohte, ``Computing with spiking neuron networks,''
  \emph{Handbook of natural computing}, pp. 335--376, 2012.

\bibitem{shrestha2018slayer}
S.~B. Shrestha and G.~Orchard, ``Slayer: Spike layer error reassignment in
  time,'' in \emph{Advances in Neural Information Processing Systems}, 2018,
  pp. 1412--1421.

\bibitem{neftci2019surrogate}
E.~O. Neftci, H.~Mostafa, and F.~Zenke, ``Surrogate gradient learning in
  spiking neural networks,'' 2019.

\bibitem{diehl2015unsupervised}
P.~U. Diehl and M.~Cook, ``Unsupervised learning of digit recognition using
  spike-timing-dependent plasticity,'' \emph{Frontiers in computational
  neuroscience}, vol.~9, p.~99, 2015.

\bibitem{perkel1968neural}
D.~H. Perkel and T.~H. Bullock, ``Neural coding.'' \emph{Neurosciences Research
  Program Bulletin}, 1968.

\bibitem{dayan2001theoretical}
P.~Dayan and L.~F. Abbott, \emph{Theoretical neuroscience: computational and
  mathematical modeling of neural systems}.\hskip 1em plus 0.5em minus
  0.4em\relax MIT press, 2001.

\bibitem{diehl2015fast}
P.~U. Diehl, D.~Neil, J.~Binas, M.~Cook, S.-C. Liu, and M.~Pfeiffer,
  ``Fast-classifying, high-accuracy spiking deep networks through weight and
  threshold balancing,'' in \emph{2015 International Joint Conference on Neural
  Networks (IJCNN)}.\hskip 1em plus 0.5em minus 0.4em\relax IEEE, 2015, pp.
  1--8.

\bibitem{georgopoulos1986neuronal}
A.~P. Georgopoulos, A.~B. Schwartz, and R.~E. Kettner, ``Neuronal population
  coding of movement direction,'' \emph{Science}, vol. 233, no. 4771, pp.
  1416--1419, 1986.

\bibitem{gerstner2014neuronal}
W.~Gerstner, W.~M. Kistler, R.~Naud, and L.~Paninski, \emph{Neuronal dynamics:
  From single neurons to networks and models of cognition}.\hskip 1em plus
  0.5em minus 0.4em\relax Cambridge University Press, 2014.

\bibitem{greschner2006complex}
M.~Greschner, A.~Thiel, J.~Kretzberg, and J.~Ammermuller, ``Complex spike-event
  pattern of transient on-off retinal ganglion cells,'' \emph{Journal of
  Neurophysiology}, vol.~96, no.~6, pp. 2845--2856, 2006.

\bibitem{panzeri2001unified}
S.~Panzeri and S.~R. Schultz, ``A unified approach to the study of temporal,
  correlational, and rate coding,'' \emph{Neural Computation}, vol.~13, no.~6,
  pp. 1311--1349, 2001.

\bibitem{rieke1999spikes}
\BIBentryALTinterwordspacing
F.~Rieke, \emph{Spikes: Exploring the Neural Code}, ser. A Bradford book.\hskip
  1em plus 0.5em minus 0.4em\relax MIT Press, 1999. [Online]. Available:
  \url{https://books.google.co.uk/books?id=7Vb9huusbcIC}
\BIBentrySTDinterwordspacing

\bibitem{zemel1998probabilistic}
R.~S. Zemel, P.~Dayan, and A.~Pouget, ``Probabilistic interpretation of
  population codes,'' \emph{Neural computation}, vol.~10, no.~2, pp. 403--430,
  1998.

\bibitem{kuncheva2014combining}
\BIBentryALTinterwordspacing
L.~Kuncheva, \emph{Combining Pattern Classifiers: Methods and
  Algorithms}.\hskip 1em plus 0.5em minus 0.4em\relax Wiley, 2014. [Online].
  Available: \url{https://books.google.co.uk/books?id=RtRLBAAAQBAJ}
\BIBentrySTDinterwordspacing

\bibitem{polikar2006ensemble}
R.~Polikar, ``Ensemble based systems in decision making,'' \emph{IEEE Circuits
  and systems magazine}, vol.~6, no.~3, pp. 21--45, 2006.

\bibitem{heskes1998selecting}
T.~Heskes, ``Selecting weighting factors in logarithmic opinion pools,'' in
  \emph{Advances in neural information processing systems}, 1998, pp. 266--272.

\bibitem{miller1999critic}
D.~J. Miller and L.~Yan, ``Critic-driven ensemble classification,'' \emph{IEEE
  Transactions on Signal Processing}, vol.~47, no.~10, pp. 2833--2844, 1999.

\bibitem{webb2019joint}
A.~M. Webb, C.~Reynolds, D.-A. Iliescu, H.~Reeve, M.~Luj{\'a}n, and G.~Brown,
  ``Joint training of neural network ensembles,'' \emph{arXiv preprint
  arXiv:1902.04422}, 2019.

\bibitem{brown2004diversity}
G.~Brown, ``Diversity in neural network ensembles,'' Ph.D. dissertation,
  Citeseer, 2004.

\bibitem{krogh1995neural}
A.~Krogh and J.~Vedelsby, ``Neural network ensembles, cross validation, and
  active learning,'' in \emph{Advances in neural information processing
  systems}, 1995, pp. 231--238.

\bibitem{kozdon2017wide}
K.~Kozdon and P.~Bentley, ``Wide learning: Using an ensemble of
  biologically-plausible spiking neural networks for unsupervised parallel
  classification of spatio-temporal patterns,'' in \emph{2017 IEEE Symposium
  Series on Computational Intelligence (SSCI)}.\hskip 1em plus 0.5em minus
  0.4em\relax IEEE, 2017, pp. 1--8.

\bibitem{stein1965theoretical}
R.~B. Stein, ``A theoretical analysis of neuronal variability,''
  \emph{Biophysical Journal}, vol.~5, no.~2, pp. 173--194, 1965.

\bibitem{abbott1999lapicque}
L.~F. Abbott, ``Lapicque’s introduction of the integrate-and-fire model
  neuron (1907),'' \emph{Brain research bulletin}, vol.~50, no. 5-6, pp.
  303--304, 1999.

\bibitem{izhikevich2003simple}
E.~M. Izhikevich, ``Simple model of spiking neurons,'' \emph{IEEE Transactions
  on neural networks}, vol.~14, no.~6, pp. 1569--1572, 2003.

\bibitem{casti2008simple}
A.~Casti, F.~Hayot, Y.~Xiao, and E.~Kaplan, ``A simple model of retina-lgn
  transmission,'' \emph{Journal of computational neuroscience}, vol.~24, no.~2,
  pp. 235--252, 2008.

\bibitem{pfeiffer2018deep}
M.~Pfeiffer and T.~Pfeil, ``Deep learning with spiking neurons: opportunities
  and challenges,'' \emph{Frontiers in neuroscience}, vol.~12, 2018.

\bibitem{hebb1949organization}
D.~O. Hebb, \emph{The organization of behavior: A neuropsychological
  theory}.\hskip 1em plus 0.5em minus 0.4em\relax Psychology Press, 1949.

\bibitem{markram1997regulation}
H.~Markram, J.~L{\"u}bke, M.~Frotscher, and B.~Sakmann, ``Regulation of
  synaptic efficacy by coincidence of postsynaptic aps and epsps,''
  \emph{Science}, vol. 275, no. 5297, pp. 213--215, 1997.

\bibitem{masquelier2009competitive}
T.~Masquelier, R.~Guyonneau, and S.~J. Thorpe, ``Competitive stdp-based spike
  pattern learning,'' \emph{Neural computation}, vol.~21, no.~5, pp.
  1259--1276, 2009.

\bibitem{daley2007introduction}
D.~J. Daley and D.~Vere-Jones, \emph{An introduction to the theory of point
  processes: volume II: general theory and structure}.\hskip 1em plus 0.5em
  minus 0.4em\relax Springer Science \& Business Media, 2007.

\bibitem{kass2014analysis}
R.~E. Kass, U.~T. Eden, and E.~N. Brown, \emph{Analysis of neural data}.\hskip
  1em plus 0.5em minus 0.4em\relax Springer, 2014, vol. 491.

\bibitem{liu2016benchmarking}
Q.~Liu, G.~Pineda-Garc{\'\i}a, E.~Stromatias, T.~Serrano-Gotarredona, and S.~B.
  Furber, ``Benchmarking spike-based visual recognition: a dataset and
  evaluation,'' \emph{Frontiers in neuroscience}, vol.~10, p. 496, 2016.

\bibitem{panda2017ensemblesnn}
P.~Panda, G.~Srinivasan, and K.~Roy, ``Ensemblesnn: distributed assistive stdp
  learning for energy-efficient recognition in spiking neural networks,'' in
  \emph{2017 International Joint Conference on Neural Networks (IJCNN)}.\hskip
  1em plus 0.5em minus 0.4em\relax IEEE, 2017, pp. 2629--2635.

\bibitem{srinivasan2018spilinc}
G.~Srinivasan, P.~Panda, and K.~Roy, ``Spilinc: spiking liquid-ensemble
  computing for unsupervised speech and image recognition,'' \emph{Frontiers in
  neuroscience}, vol.~12, p. 524, 2018.

\bibitem{shadlen1998variable}
M.~N. Shadlen and W.~T. Newsome, ``The variable discharge of cortical neurons:
  implications for connectivity, computation, and information coding,''
  \emph{Journal of neuroscience}, vol.~18, no.~10, pp. 3870--3896, 1998.

\bibitem{bichler2012extraction}
O.~Bichler, D.~Querlioz, S.~J. Thorpe, J.-P. Bourgoin, and C.~Gamrat,
  ``Extraction of temporally correlated features from dynamic vision sensors
  with spike-timing-dependent plasticity,'' \emph{Neural Networks}, vol.~32,
  pp. 339--348, 2012.

\bibitem{querlioz2011simulation}
D.~Querlioz, O.~Bichler, and C.~Gamrat, ``Simulation of a memristor-based
  spiking neural network immune to device variations,'' in \emph{The 2011
  International Joint Conference on Neural Networks}.\hskip 1em plus 0.5em
  minus 0.4em\relax IEEE, 2011, pp. 1775--1781.

\bibitem{sherrington1952integrative}
C.~Sherrington, \emph{The integrative action of the nervous system}.\hskip 1em
  plus 0.5em minus 0.4em\relax CUP Archive, 1952.

\bibitem{buzsaki2010neural}
G.~Buzs{\'a}ki, ``Neural syntax: cell assemblies, synapsembles, and readers,''
  \emph{Neuron}, vol.~68, no.~3, pp. 362--385, 2010.

\bibitem{deadwyler1997significance}
S.~A. Deadwyler and R.~E. Hampson, ``The significance of neural ensemble codes
  during behavior and cognition,'' \emph{Annual review of neuroscience},
  vol.~20, no.~1, pp. 217--244, 1997.

\bibitem{ince2013neural}
R.~A. Ince, S.~Panzeri, and C.~Kayser, ``Neural codes formed by small and
  temporally precise populations in auditory cortex,'' \emph{Journal of
  Neuroscience}, vol.~33, no.~46, pp. 18\,277--18\,287, 2013.

\bibitem{sakurai2018multiple}
Y.~Sakurai, Y.~Osako, Y.~Tanisumi, E.~Ishihara, J.~Hirokawa, and H.~Manabe,
  ``Multiple approaches to the investigation of cell assembly in memory
  research—present and future,'' \emph{Frontiers in systems neuroscience},
  vol.~12, 2018.

\bibitem{burgess1994model}
N.~Burgess, M.~Recce, and J.~O'Keefe, ``A model of hippocampal function,''
  \emph{Neural networks}, vol.~7, no. 6-7, pp. 1065--1081, 1994.

\bibitem{georgopoulos1982relations}
A.~P. Georgopoulos, J.~F. Kalaska, R.~Caminiti, and J.~T. Massey, ``On the
  relations between the direction of two-dimensional arm movements and cell
  discharge in primate motor cortex,'' \emph{Journal of Neuroscience}, vol.~2,
  no.~11, pp. 1527--1537, 1982.

\bibitem{maunsell1983functional}
J.~H. Maunsell and D.~C. Van~Essen, ``Functional properties of neurons in
  middle temporal visual area of the macaque monkey. i. selectivity for
  stimulus direction, speed, and orientation,'' \emph{Journal of
  neurophysiology}, vol.~49, no.~5, pp. 1127--1147, 1983.

\bibitem{georgopoulos1988primate}
A.~P. Georgopoulos, R.~E. Kettner, and A.~B. Schwartz, ``Primate motor cortex
  and free arm movements to visual targets in three-dimensional space. ii.
  coding of the direction of movement by a neuronal population,'' \emph{Journal
  of Neuroscience}, vol.~8, no.~8, pp. 2928--2937, 1988.

\bibitem{salinas1994vector}
E.~Salinas and L.~Abbott, ``Vector reconstruction from firing rates,''
  \emph{Journal of computational neuroscience}, vol.~1, no. 1-2, pp. 89--107,
  1994.

\bibitem{lecun-mnisthandwrittendigit-2010}
\BIBentryALTinterwordspacing
Y.~LeCun and C.~Cortes, ``{MNIST} handwritten digit database,''
  http://yann.lecun.com/exdb/mnist/, 2010. [Online]. Available:
  \url{http://yann.lecun.com/exdb/mnist/}
\BIBentrySTDinterwordspacing

\bibitem{orchard2015converting}
G.~Orchard, A.~Jayawant, G.~K. Cohen, and N.~Thakor, ``Converting static image
  datasets to spiking neuromorphic datasets using saccades,'' \emph{Frontiers
  in neuroscience}, vol.~9, p. 437, 2015.

\bibitem{bi1998synaptic}
G.-q. Bi and M.-m. Poo, ``Synaptic modifications in cultured hippocampal
  neurons: dependence on spike timing, synaptic strength, and postsynaptic cell
  type,'' \emph{Journal of neuroscience}, vol.~18, no.~24, pp.
  10\,464--10\,472, 1998.

\bibitem{Stimberg2019}
M.~Stimberg, R.~Brette, and D.~F. Goodman, ``Brian 2, an intuitive and
  efficient neural simulator,'' \emph{eLife}, vol.~8, p. e47314, Aug. 2019.

\bibitem{shim2016unsupervised}
Y.~Shim, A.~Philippides, K.~Staras, and P.~Husbands, ``Unsupervised learning in
  an ensemble of spiking neural networks mediated by itdp,'' \emph{PLoS
  computational biology}, vol.~12, no.~10, p. e1005137, 2016.

\bibitem{wu2019direct}
Y.~Wu, L.~Deng, G.~Li, J.~Zhu, Y.~Xie, and L.~Shi, ``Direct training for
  spiking neural networks: Faster, larger, better,'' in \emph{Proceedings of
  the AAAI Conference on Artificial Intelligence}, vol.~33, 2019, pp.
  1311--1318.

\bibitem{bohte2000spikeprop}
S.~M. Bohte, J.~N. Kok, and J.~A. La~Poutr{\'e}, ``Spikeprop: backpropagation
  for networks of spiking neurons.'' in \emph{ESANN}, vol.~48, 2000, pp.
  17--37.

\bibitem{mostafa2017supervised}
H.~Mostafa, ``Supervised learning based on temporal coding in spiking neural
  networks,'' \emph{IEEE transactions on neural networks and learning systems},
  vol.~29, no.~7, pp. 3227--3235, 2017.

\bibitem{anderson1994neurobiological}
C.~H. Anderson and D.~C. Van~Essen, ``Neurobiological computational systems,''
  \emph{IEEE World Congress on Computational Intelligence}, 1994.

\bibitem{rathi2018stdp}
N.~Rathi and K.~Roy, ``Stdp-based unsupervised multimodal learning with
  cross-modal processing in spiking neural network,'' \emph{IEEE Transactions
  on Emerging Topics in Computational Intelligence}, 2018.

\bibitem{neculae2020ensemble}
G.~Neculae, ``Ensemble learning for spiking neural networks,'' Ph.D.
  dissertation, The University of Manchester (United Kingdom), 2020.

\bibitem{sarle1994neural}
W.~S. Sarle, ``Neural networks and statistical models,'' \emph{Proceedings of
  the Nineteenth Annual SAS Users Group International Conference}, 1994.

\bibitem{shlens2014notes}
J.~Shlens, ``Notes on generalized linear models of neurons,'' \emph{arXiv
  preprint arXiv:1404.1999}, 2014.

\bibitem{truccolo2005point}
W.~Truccolo, U.~T. Eden, M.~R. Fellows, J.~P. Donoghue, and E.~N. Brown, ``A
  point process framework for relating neural spiking activity to spiking
  history, neural ensemble, and extrinsic covariate effects,'' \emph{Journal of
  neurophysiology}, vol.~93, no.~2, pp. 1074--1089, 2005.

\bibitem{heeger2000poisson}
D.~Heeger, ``Poisson model of spike generation,'' \emph{Handout, University of
  Standford}, vol.~5, pp. 1--13, 2000.

\bibitem{perkel1967neuronal}
D.~H. Perkel, G.~L. Gerstein, and G.~P. Moore, ``Neuronal spike trains and
  stochastic point processes: I. the single spike train,'' \emph{Biophysical
  journal}, vol.~7, no.~4, pp. 391--418, 1967.

\bibitem{battiti1994democracy}
R.~Battiti and A.~M. Colla, ``Democracy in neural nets: Voting schemes for
  classification,'' \emph{Neural Networks}, vol.~7, no.~4, pp. 691--707, 1994.

\bibitem{kittler1998combining}
J.~Kittler, M.~Hatef, R.~P. Duin, and J.~Matas, ``On combining classifiers,''
  \emph{IEEE transactions on pattern analysis and machine intelligence},
  vol.~20, no.~3, pp. 226--239, 1998.

\bibitem{alkoot1999experimental}
F.~M. Alkoot and J.~Kittler, ``Experimental evaluation of expert fusion
  strategies,'' \emph{Pattern recognition letters}, vol.~20, no. 11-13, pp.
  1361--1369, 1999.

\bibitem{kuncheva2002theoretical}
L.~I. Kuncheva, ``A theoretical study on six classifier fusion strategies,''
  \emph{IEEE Transactions on pattern analysis and machine intelligence},
  vol.~24, no.~2, pp. 281--286, 2002.

\bibitem{fang2020exploiting}
H.~Fang, A.~Shrestha, Z.~Zhao, and Q.~Qiu, ``Exploiting neuron and synapse
  filter dynamics in spatial temporal learning of deep spiking neural
  network,'' \emph{arXiv preprint arXiv:2003.02944}, 2020.

\bibitem{cheng2020finite}
X.~Cheng, T.~Zhang, S.~Jia, and B.~Xu, ``Finite meta-dynamic neurons in spiking
  neural networks for spatio-temporal learning,'' \emph{arXiv preprint
  arXiv:2010.03140}, 2020.

\bibitem{kugele2020efficient}
A.~Kugele, T.~Pfeil, M.~Pfeiffer, and E.~Chicca, ``Efficient processing of
  spatio-temporal data streams with spiking neural networks,'' \emph{Frontiers
  in Neuroscience}, vol.~14, p. 439, 2020.

\end{thebibliography}
\end{document}